\documentclass{article}
\usepackage[preprint]{neurips_2026}

\PassOptionsToPackage{numbers,sort&compress}{natbib}
\usepackage{amsmath}
\usepackage{wrapfig}
\usepackage{graphicx}
\usepackage{neurips_2026}
\usepackage{tikz}
\usetikzlibrary{arrows.meta,calc}
\usetikzlibrary{decorations.pathreplacing}

\usepackage{threeparttable}
\usepackage[table,dvipsnames]{xcolor}
\usepackage[titletoc,title]{appendix}
\usepackage{subcaption}

\usepackage{amsmath}
\usepackage{bm}

\newcommand{\bestsci}[3]{\ensuremath{\boldsymbol{#1{\times}10^{#2}}\pmsem{#3}}}
\newcommand{\secondsci}[3]{\ensuremath{\underline{#1{\times}10^{#2}}\pmsem{#3}}}

\newcommand{\bestvalerr}[2]{\ensuremath{\boldsymbol{#1}\pmsem{#2}}}
\newcommand{\secondvalerr}[2]{\ensuremath{\underline{#1}\pmsem{#2}}}

\newcommand{\bestcap}[1]{\textbf{#1}}
\newcommand{\secondcap}[1]{\underline{#1}}

\newcommand{\sci}[2]{\ensuremath{#1{\times}10^{#2}}}
\newcommand{\pmsem}[1]{\ensuremath{\,{\scriptstyle \pm\,#1}}}
\newcommand{\valerr}[2]{\ensuremath{#1{\scriptstyle \pm\,#2}}}

\usepackage[utf8]{inputenc} 
\usepackage[T1]{fontenc}    
\usepackage{hyperref}       
\usepackage{url}            
\usepackage{booktabs}       
\usepackage{amsfonts}       
\usepackage{nicefrac}       
\usepackage{microtype}      
\usepackage{xcolor}         
\usepackage{multirow}
\usepackage{threeparttable}
\usepackage[utf8]{inputenc}
\usepackage[T1]{fontenc}
\title{CORE: Cayley-Parameterized Rotating Embeddings for Object Discovery}

\author{%
  Despoina Touska\\
  Faculty of Information, Media and \\
  Electrical Engineering\\
   TH K\"oln, Cologne, Germany\\
  \texttt{despoina.touska@th-koeln.de} \\
  \And
  Pascal Cerfontaine\\
  Faculty of Information, Media and \\
  Electrical Engineering\\
   TH K\"oln, Cologne, Germany\\
  \texttt{pascal.cerfontaine@th-koeln.de} \\
}

\begin{document}

\maketitle

\begin{abstract}
Inspired by the brain's use of neural synchrony for object binding, synchrony-based neural networks provide a promising framework for Object-Centric Learning (OCL). Recent architectures such as Rotating Features (RF) extend synchrony-based representations beyond 2D complex-valued activations by modeling features as \(n\)-dimensional vectors on a hypersphere. However, RF induces phase shifts through additive latent biases. From a geometric perspective, these biases correspond to translations in feature space, which can alter feature magnitudes and entangle phase-based object affiliation with magnitude-based semantic strength.
We propose CORE (Cayley-Orthogonal Rotating Embeddings), a geometry-preserving extension of RF that replaces additive biases with learned orthogonal transformations along the rotation dimension. CORE uses a rank-2 Cayley parameterization to learn per-channel phase updates that preserve feature norms. By rotating features directly in the rotation space, CORE enables object-affiliation updates without altering semantic magnitude, providing a principled orthogonal mechanism for unsupervised object discovery. Evaluations on both synthetic and real-image datasets validate the effectiveness of this approach.

\end{abstract}

\section{Introduction}
\label{introduction}
Reasoning across time and space is an essential requirement for any intelligent system. To reason effectively, systems must have mechanisms that can hold and manipulate information in memory over extended periods. While standard neural network architectures often struggle with information degradation over time and space~\cite{vaswani2017attention}, nature has long solved this problem through a universal mechanism: wave dynamics. From the propagation of electromagnetic waves across the universe to the complex neural oscillations within our brain~\cite{muller2018cortical,milner1974model}, waves serve as a robust medium for both transmitting and preserving information. In this sense, waves are not merely a phenomenon of physics, but a general computational principle: they propagate information while maintaining coherent structure over space and time. This makes wave-like dynamics an appealing foundation for building artificial systems that must reason, remember, and bind information across long spatiotemporal scales.

Inspired by these natural phenomena, the machine learning community has developed architectures with more structured forms of memory. For example, Wave-RNN~\cite{keller2023traveling}, a minimal recurrent neural network designed to produce traveling waves in its hidden state, demonstrated that wave-like dynamics can encode recent history more effectively than wave-free counterparts. In parallel, researchers have increasingly explored neural networks with complex-valued representations~\cite{gopalakrishnan2024recurrent,stanic2023contrastive,lowe2022complex}, in which units are characterized by both amplitude and phase. Notable examples include models~\cite{miyato2024artificial,muzellec2025enhancing} based on Kuramoto oscillators~\cite{acebron2005kuramoto,rodrigues2016kuramoto}, where coupled units dynamically adjust their phases to synchronize. Such synchronization-based mechanisms have been applied to tasks such as object discovery~\cite{lowe2022complex,stanic2023contrastive,gopalakrishnan2024recurrent,lowe2023rotating}, adversarial robustness~\cite{miyato2024artificial}, and reasoning~\cite{miyato2024artificial}.

\begin{figure}[t]
\centering
\begin{tikzpicture}[
    scale=1.0,
    >=Stealth,
    vector/.style={very thick, -{Stealth[length=3.2mm,width=2.4mm]}},
    manifold/.style={dashed, thick, gray!60},
    panelred/.style={rounded corners=8pt, draw=red!70!black, thick},
    panelgreen/.style={rounded corners=8pt, draw=green!50!black, thick},
    eqboxred/.style={rounded corners=5pt, draw=red!50!black, fill=red!5, inner sep=4pt},
    eqboxgreen/.style={rounded corners=5pt, draw=green!50!black, fill=green!5, inner sep=4pt}
]

\begin{scope}[shift={(0,0)}]

    \draw[panelred] (-2.35, -2.55) rectangle (4.35, 4.0);

    \node[font=\bfseries\large, text=red!75!black, align=center] at (1.0, 3.7)
    {RF: additive phase shift};

    \node[text=red!75!black, font=\normalsize, align=center, text width=5.5cm] at (1.0, 3.2)
    {$\textcolor{blue!80!black}{f_{\mathbf w}(\mathbf z_{\mathrm{in}})}
    \;\longmapsto\;
    \textcolor{blue!80!black}{f_{\mathbf w}(\mathbf z_{\mathrm{in}})}+\textcolor{red!80!black}{\mathbf b}$};

    \begin{scope}[yshift=0.35cm]
        \draw[->, gray!60] (-1.9,0) -- (3.5,0) node[right, black] {$e_1$};
        \draw[->, gray!60] (0,-1.8) -- (0,2.25) node[above, black] {$e_2$};
        \node[below left, xshift=-1pt, yshift=-1pt] at (0,0) {$0$};

        \draw[manifold] (0,0) circle (1.30cm);

        \coordinate (PsiL) at (0.75, 1.06);
        \draw[vector, draw=blue!80!black] (0,0) -- (PsiL)
            node[midway, above left=-2pt, font=\small] {$\|\psi\|_2$};
        \node[text=blue!80!black, above=2pt] at (PsiL) {$\psi$};

        \coordinate (BiasShift) at (1.35, -0.35);
        \coordinate (BiasEnd) at (2.10, 0.71);
        \draw[vector, draw=orange!90!black] (PsiL) -- (BiasEnd)
            node[midway, above=1pt] {$\mathbf b$};

        \draw[vector, draw=red!80!black] (0,0) -- (BiasEnd)
            node[midway, below right=-2pt, font=\small] {$\|\psi+\mathbf b\|_2$};
        \node[text=red!80!black, right=2pt] at (BiasEnd) {$\psi+\mathbf b$};

        \draw[manifold, opacity=0.55] (BiasShift) circle (1.30cm);
    \end{scope}

    \node[eqboxred] at (1.0,-2.05)
    {$\|\psi+\mathbf b\|_2 \neq \|\psi\|_2$};

\end{scope}

\begin{scope}[shift={(7.2,0)}]

    \draw[panelgreen] (-2.35, -2.55) rectangle (4.35, 4.0);

    \node[font=\bfseries\large, text=green!50!black, align=center] at (1.0, 3.7)
    {CORE: orthogonal phase update};

    \node[text=green!50!black, font=\normalsize, align=center, text width=5.5cm] at (1.0, 3.2)
    {$\textcolor{blue!80!black}{f_{\mathbf w}(\mathbf z_{\mathrm{in}})}
    \;\longmapsto\;
    \textcolor{green!60!black}{Q}\,\textcolor{blue!80!black}{f_{\mathbf w}(\mathbf z_{\mathrm{in}})}$};

    \begin{scope}[yshift=0.35cm, xshift=1.0cm]
        \draw[->, gray!60] (-2.9,0) -- (2.9,0) node[right, black] {$e_1$};
        \draw[->, gray!60] (0,-1.8) -- (0,2.25) node[above, black] {$e_2$};
        \node[below left, xshift=-1pt, yshift=-1pt] at (0,0) {$0$};

        \draw[manifold] (0,0) circle (1.30cm);

        \coordinate (PsiR) at (0.75, 1.06);
        \draw[vector, draw=blue!80!black] (0,0) -- (PsiR)
            node[midway, below right=-2pt, font=\small] {$\|\psi\|_2$};
        \node[text=blue!80!black, right=2pt] at (PsiR) {$\psi$};

        \coordinate (PsiRot) at (-1.06, 0.75);
        \draw[vector, draw=green!60!black] (0,0) -- (PsiRot)
            node[midway, below left=-2pt, font=\small] {$\|Q\psi\|_2$};
        \node[text=green!60!black, left=2pt] at (PsiRot) {$Q\psi$};

        \draw[-{Stealth[length=3mm,width=2mm]}, thick, draw=green!60!black]
            (55:1.60) arc[start angle=55, end angle=145, radius=1.60];
    \end{scope}

    \node[eqboxgreen] at (1.0,-2.05)
    {$\|Q\psi\|_2 = \|\psi\|_2$};

\end{scope}

\end{tikzpicture}

\caption{Comparison of phase-update mechanisms in rotating-feature architectures.
\textbf{Left:} RF applies an additive bias, which acts as a Euclidean translation and can change the feature norm.
\textbf{Right:} CORE applies an orthogonal transformation \(Q\), which updates phase while preserving the norm.
}
\label{fig:geometric_flaw_vs_fix}
\end{figure}
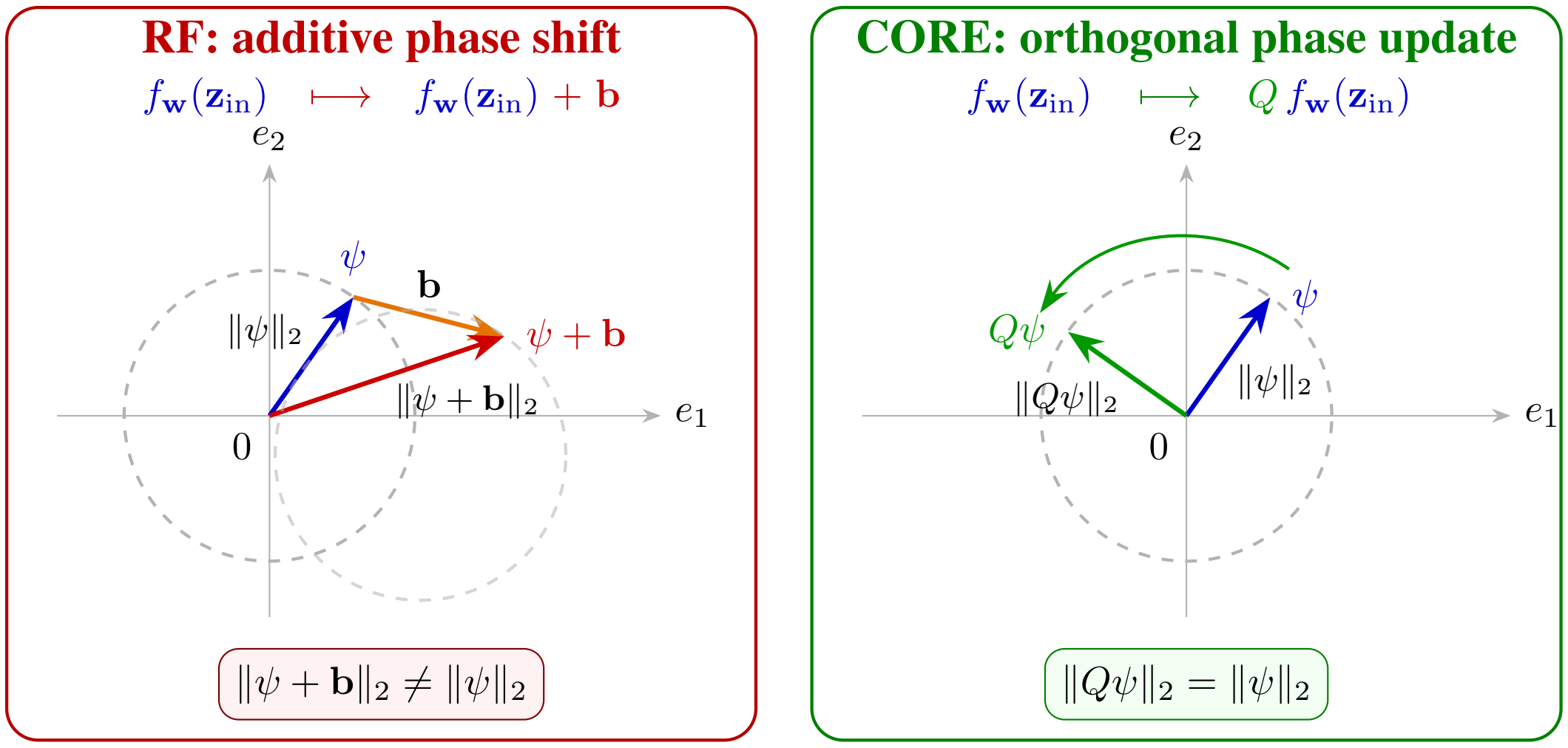

This synchronization is not merely a mathematical curiosity; it is deeply rooted in cognitive neuroscience~\cite{koban2019we,orsucci2013prosody}. A prevailing hypothesis for how the brain solves the "binding problem"~\cite{roskies1999binding}—the process of integrating disparate sensory features into a coherent, object-centric representation—is through the temporal synchronization of neural oscillations~\cite{engel1997role,fries2005mechanism,fries2015rhythms,palmigiano2017flexible}. When distinct neurons processing parts of the same object align their phases, they "bind" those features together~\cite{singer2009distributed,singer1995visual}.

While phase-synchronization offers a compelling framework for Object-Centric Learning (OCL), implementing this effectively in deep neural networks remains a geometric challenge. 
State-of-the-art synchrony-based models, such as Complex-valued Autoencoders (CAEs)~\cite{lowe2022complex}, address this by processing complex-valued activations using real-valued weights shared across the real and imaginary components. In this paradigm, magnitude encodes feature presence, while phase encodes object affiliation. These models inherently leverage the constructive and destructive interference that occurs when adding complex activations, naturally encouraging phase alignment among features of the same object and phase separation among distinct objects. Recently, Rotating Features (RF)~\cite{lowe2023rotating} expanded this approach by replacing 2D complex representations with $n$-dimensional vector-valued activations that rotate on a hypersphere, significantly enhancing representational capacity. 

Despite this advancement, the original RF~\cite{lowe2023rotating} architecture relied on additive latent biases to induce these phase shifts. Geometrically, additive shifts act as translations that inherently distort feature magnitudes (the $\ell_2$ norm) and entangle representations. To achieve true feature rotation, a network must be able to adjust the "phase" of an embedding without destroying its underlying semantic magnitude. In this work, we propose a novel approach to synchrony-based OCL by introducing CORE: Cayley-Orthogonal Rotating Embeddings. Rather than relying on geometrically destructive additive biases, we parameterize feature synchronization as a strict geometric rotation using a low-rank Cayley transform~\cite{qiu2023controlling,jauch2020random,trockman2021orthogonalizing}. By learning an optimal 2D plane of rotation for each feature channel, our network allows features to smoothly "spin" and align (see Figure~\ref{fig:geometric_flaw_vs_fix}). Crucially, because these operations are strictly orthogonal, they preserve semantic magnitudes while leveraging spatial competition to group synchronized representations into distinct objects. 
Experimentally, we show that CORE improves the performance of synchrony-based models on object discovery tasks in synthetic RGB images, while also demonstrating its applicability to real-world images.

\section{Background}
\label{background}
\paragraph{Vector-valued representations and phase cancellation}
In synchrony-based architectures~\citep{lowe2023rotating,lowe2022complex,stanic2023contrastive} features are represented as vectors rather than scalars. In this formulation, a vector's magnitude encodes the presence or strength of a feature, while its orientation (phase) signifies its affiliation. Objects are natively bound via constructive interference of similarly oriented vectors. However, relying purely on linear vector addition introduces a critical geometric vulnerability: destructive interference. If two active features possess similar magnitudes but perfectly out-of-phase orientations, their linear average collapses toward zero. This cancellation effectively erases strong feature evidence from the latent space simply due to clashing object affiliations.

\begin{wrapfigure}{r}{0.4\linewidth}
    \vspace{-12pt}
    \centering
    \includegraphics[width=0.98\linewidth]{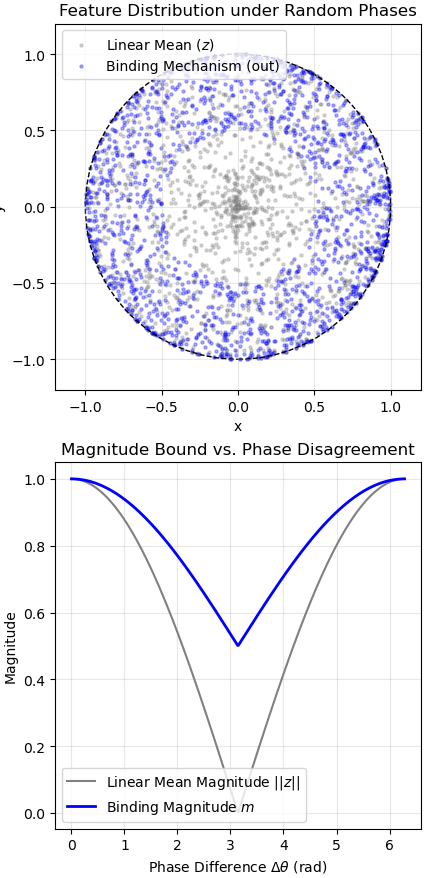}
    \caption{\small
    Toy binding example. \textbf{(Top)} Feature distribution for random input phases. Linear averaging (gray) can collapse representations toward the origin, whereas binding (blue) maintains a lower-bounded magnitude ($m \ge 0.5$), preserving feature evidence under destructive interference.
\textbf{(Bottom)} Output magnitude as a function of phase difference of two inputs. With the $\chi$ term contributing to the gated magnitude, out-of-phase excitation cannot cancel completely.} 
\label{fig:binding_mechanism
    }
    \label{fig:binding_mechanism}
    \vspace{10pt}
\end{wrapfigure}

\paragraph{Binding mechanism}
To decouple feature presence from phase alignment, we employ a gating strategy introduced by \citet{reichert2013neuronal} and subsequently adopted in synchrony-based representation learning~\citep{lowe2023rotating,lowe2022complex,stanic2023contrastive, touskaorthorf}. We first lift a standard feature vector $\mathbf z\in\mathbb{R}^d$ to a vector-valued representation $\mathbf z_{\mathrm{rotating}}\in\mathbb{R}^{n\times d}$. The magnitude of each feature vector,
$\mathbf m=\|\mathbf z_{\mathrm{rotating}}\|_2\in\mathbb{R}^d$, computed over the $n$ rotating dimensions, plays the role of a conventional scalar activation, while the orientation encodes grouping information.

Given a layer $f_{\mathbf w}$, the vector-valued input $\mathbf z_{\mathrm{in}}\in\mathbb{R}^{n\times d_{\mathrm{in}}}$ is transformed using weights $\mathbf w$ shared across the rotating dimensions, together with an additive bias offset $\mathbf b$. This produces the vector-valued activation:
\begin{equation}
\boldsymbol{\psi}
=
f_{\mathbf w}(\mathbf z_{\mathrm{in}}) + \mathbf b
\in \mathbb{R}^{n\times d_{\mathrm{out}}}.
\label{eq:conv}
\end{equation}
If the output magnitude were determined solely by the vector-valued response, strongly active features with opposing orientations could cancel through destructive interference. To prevent this, the binding mechanism computes a complementary magnitude-only branch by applying the same layer to the input magnitudes, and then defines the gated magnitude as the average of the vector-response norm and this magnitude-only baseline:
\begin{equation}
  {\mathbf m}_\mathrm{bind} = 0.5 \cdot || \boldsymbol{\psi}||_{2} + 0.5 \cdot \boldsymbol{\chi}\in \mathbb{R}^{d_\mathrm{out}},
  \label{eq:avg}
\end{equation}

where:
\begin{equation}
  \boldsymbol{\chi} = f_{{\mathbf w}} (|| {\mathbf z}_\mathrm{in}||_{2}) \in \mathbb{R}^{d_\mathrm{out}}.
\end{equation}

Finally, this stabilized magnitude is used to rescale the normalized vector-valued response,
\begin{equation}
\mathbf z_{\mathrm{out}}
=
\mathbf m_{\mathrm{bind}}
\odot
\frac{\boldsymbol{\psi}}{\|\boldsymbol{\psi}\|_2}.
\end{equation}

This operation preserves the orientation of $\boldsymbol{\psi}$, and therefore the phase-based interaction between features, while ensuring that out-of-phase excitations do not cancel completely. As shown by~\citet{lowe2023rotating}, this gating step is essential for forming object-centric representations. Finally, the model is trained by reconstructing the input image from the magnitude of the final-layer activations across the rotation dimension and minimizing the Mean Squared Error (MSE) between the reconstruction and the original image.

\paragraph{Toy example: binding with the mean operator}
\label{sec:toy_binding}

To illustrate how magnitude gating mitigates collapse, consider a minimal 2D interaction in which $f_{\mathbf w}$ is replaced by the mean of two vectors, $\psi=(v_1+v_2)/2$. In parallel, we compute the phase-invariant branch from the input norms, $\chi=(\|v_1\|_2+\|v_2\|_2)/2$. The gated magnitude is then $m=(\|\psi\|_2+\chi)/2$, and the output is given by $\mathrm{out}=m \odot z/\max(\|\psi\|_2,\epsilon)$, where $\epsilon$ avoids division by zero when the mean direction is undefined.
This construction lower-bounds the gated magnitude. Since:
\begin{equation}
m
=
\frac{\|\psi\|_2+\chi}{2}
\geq
\frac{\|v_1\|_2+\|v_2\|_2}{4},
\end{equation}
strong inputs cannot be fully suppressed by destructive interference alone. For unit-norm inputs, $\chi=1$ and therefore $m\geq 0.5$. As shown in Figure~\ref{fig:binding_mechanism}, this lower bound creates a geometric exclusion region around the origin: out-of-phase features are suppressed to reflect disagreement, but their underlying excitation never cancels out completely.

\section{Method}
\label{method}
While RF~\cite{lowe2023rotating} binds objects effectively, its additive biases on rotating activations introduce a geometric inconsistency: they translate features in latent space, altering magnitudes and entangling direction with scale. We propose Cayley-Orthogonal Rotating Embeddings (CORE), which replaces additive shifts with learned orthogonal transformations that act across the rotation dimension. We first analyze the limitation of additive shifts, then formulate feature synchronization as a norm-preserving orthogonal transformation, and finally use a rank-2 Cayley parameterization to learn these rotations efficiently.

\begin{figure}[t]
    \centering
    \includegraphics[width=1.0\linewidth]{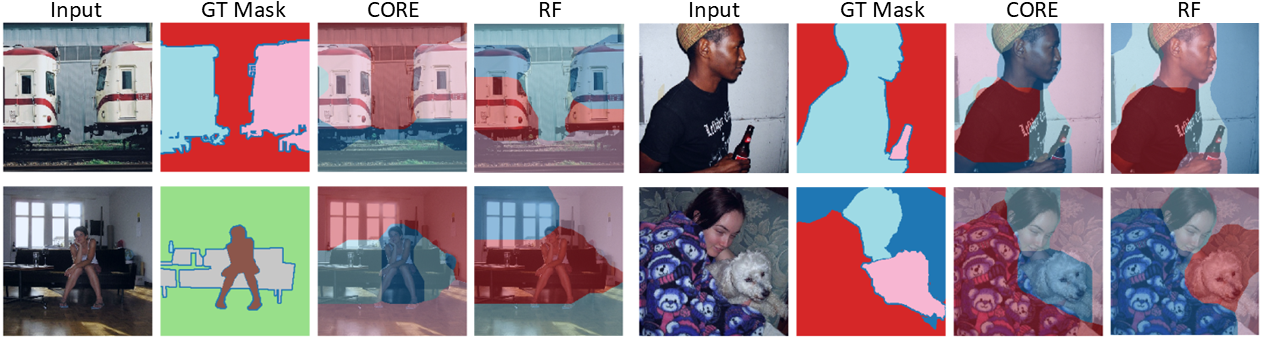}
    \caption{Qualitative comparison between CORE and RF~\cite{lowe2023rotating} on the Pascal VOC dataset~\cite{everingham2010pascal}.}
    \label{fig:pascal}
\end{figure}

\subsection{Geometric limitation of additive phase shifts}

As shown in Equation~\ref{eq:conv}, prior synchrony-based models such as RF~\cite{lowe2023rotating} break symmetry among rotating features by introducing an additive bias \(\mathbf{b}\) during the layer computation. Geometrically, this operation corresponds to a Euclidean translation in the latent feature space rather than a rotation in the phase space (see Figure~\ref{fig:geometric_flaw_vs_fix}). Consequently, it does not preserve distance to the origin and therefore generally changes the feature norm:
\begin{equation}
    \left\|
    f_{\mathbf{w}}(\mathbf{z}_{\mathrm{in}})
    +
    \mathbf{b}
    \right\|_2
    \neq
    \left\|
    f_{\mathbf{w}}(\mathbf{z}_{\mathrm{in}})
    \right\|_2 .
\end{equation}

This is problematic for rotating-feature representations, where magnitude encodes feature presence and phase encodes object affiliation. Phase synchronization should therefore preserve magnitude. Additive bias violates this principle by translating activations rather than rotating them, thereby entangling phase alignment with magnitude modulation and potentially altering semantic evidence during object assignment.

\subsection{Parameterizing orthogonal phase synchronization}

To decouple feature presence from object affiliation, phase synchronization should be parameterized as a norm-preserving transformation rather than an additive translation. We therefore introduce an orthogonal operator \(Q \in O(n)\), acting along the rotation dimension, such that
\begin{equation}
    \boldsymbol{\psi}' = Q\boldsymbol{\psi}.
\end{equation}
By definition, \(Q\) satisfies \(Q^\top Q = I\), which guarantees norm preservation, i.e., \(\|Q\boldsymbol{\psi}\|_2 = \|\boldsymbol{\psi}\|_2\). This ensures that synchronizing the phase of a rotating feature does not change its magnitude. Moreover, orthogonal transformations preserve inner products and therefore maintain relative angles and cosine similarities between rotating feature vectors. As a result, the network can update object affiliation through phase alignment while preserving both semantic strength and the geometric relationships among features.

\subsection{Cayley parameterization of orthogonal phase updates}

Directly optimizing an unconstrained matrix \(Q\) under the orthogonality constraint \(Q^\top Q = I\) is challenging with standard gradient-based training, since ordinary gradient steps do not generally remain on the orthogonal group. Enforcing this constraint through explicit projection or orthogonalization, such as QR decomposition~\cite{gander1980algorithms} or Gram--Schmidt~\cite{leon2013gram} procedures, can introduce additional computational overhead and complicate optimization.
To obtain orthogonal phase updates without explicit projection, we parameterize \(Q\) using the Cayley transform~\cite{cayley1846quelques}. Let \(S \in \mathbb{R}^{n \times n}\) be a skew-symmetric matrix satisfying \(S^\top = -S\). The Cayley transform maps \(S\) to an orthogonal matrix:
\begin{equation}
    Q = (I - S)^{-1}(I + S).
\end{equation}
Since \(S\) is skew-symmetric, the resulting matrix satisfies \(Q^\top Q = I\). Thus, the transformation preserves the norm of the rotating feature representation while remaining differentiable with respect to the parameters used to construct \(S\). In CORE, this allows phase transformations to be learned through standard backpropagation while guaranteeing orthogonality at every update.

\begin{table}[t]
\centering
\caption{Comparison of phase-update mechanisms in rotating-feature architectures.}
\label{tab:phase_update_comparison}
\renewcommand{\arraystretch}{1.8} 
\resizebox{\textwidth}{!}{
\begin{tabular}{lccc}
\toprule
\textbf{Method}
&
\textbf{RF Additive Bias}
&
\textbf{Full Cayley}
&
\textbf{CORE Rank-2 Cayley}
\\
\midrule

\textbf{Illustration}
&
\raisebox{-0.5\height}{
\begin{tikzpicture}[scale=0.35]
  \fill[gray!40] (0,0) rectangle (4,3);
  \draw[step=1, black, very thin] (0,0) grid (4,3);
  \draw[decorate,decoration={brace,amplitude=3pt,mirror},yshift=-2pt] (0,0) -- (4,0) node[midway,below=2pt] {\scriptsize $n$};
  \draw[decorate,decoration={brace,amplitude=3pt},xshift=-2pt] (0,0) -- (0,3) node[midway,left=2pt] {\scriptsize $C$};
\end{tikzpicture}
}
&
\raisebox{-0.5\height}{
\begin{tikzpicture}[scale=0.35]
  \fill[orange!50] (0,0) rectangle (4,4);
  \draw[step=1, black, very thin] (0,0) grid (4,4);
  \draw[decorate,decoration={brace,amplitude=3pt,mirror},yshift=-2pt] (0,0) -- (4,0) node[midway,below=2pt] {\scriptsize $n$};
  \draw[decorate,decoration={brace,amplitude=3pt},xshift=-2pt] (0,0) -- (0,4) node[midway,left=2pt] {\scriptsize $n$};
  \draw[decorate,decoration={brace,amplitude=4pt},xshift=-29pt] (0,0) -- (0,4) node[midway,left=4pt] {$C \times$};
\end{tikzpicture}
}
&
\raisebox{-0.5\height}{
\begin{tikzpicture}[scale=0.35]
  \draw[decorate,decoration={brace,amplitude=4pt},xshift=-29pt] (0,0) -- (0,4) node[midway,left=4pt] {$C \times$};

  \fill[orange!60] (0,0) rectangle (1,4); 
  \draw[step=1, black, very thin] (0,0) grid (1,4);
  \draw[decorate,decoration={brace,amplitude=3pt,mirror},yshift=-2pt] (0,0) -- (1,0) node[midway,below=2pt] {\scriptsize $1$};
  \draw[decorate,decoration={brace,amplitude=3pt},xshift=-2pt] (0,0) -- (0,4) node[midway,left=2pt] {\scriptsize $n$};
  
  \node at (1.8, 2) {$\times$};
  
  \begin{scope}[shift={(2.6, 1.5)}]
      \fill[gray!50] (0,0) rectangle (4,1); 
      \draw[step=1, black, very thin] (0,0) grid (4,1);
      \draw[decorate,decoration={brace,amplitude=3pt,mirror},yshift=-2pt] (0,0) -- (4,0) node[midway,below=2pt] {\scriptsize $n$};
  \end{scope}
  
  \node at (7.6, 2) {$-$};
  
  \begin{scope}[shift={(10.0, 0)}]
      \fill[gray!50] (0,0) rectangle (1,4); 
      \draw[step=1, black, very thin] (0,0) grid (1,4);
      \draw[decorate,decoration={brace,amplitude=3pt,mirror},yshift=-2pt] (0,0) -- (1,0) node[midway,below=2pt] {\scriptsize $1$};
      \draw[decorate,decoration={brace,amplitude=3pt},xshift=-2pt] (0,0) -- (0,4) node[midway,left=2pt] {\scriptsize $n$};
  \end{scope}
  
  \node at (11.8, 2) {$\times$};
  
  \begin{scope}[shift={(12.6, 1.5)}]
      \fill[orange!60] (0,0) rectangle (4,1); 
      \draw[step=1, black, very thin] (0,0) grid (4,1);
      \draw[decorate,decoration={brace,amplitude=3pt,mirror},yshift=-2pt] (0,0) -- (4,0) node[midway,below=2pt] {\scriptsize $n$};
  \end{scope}
\end{tikzpicture}
}
\\

\textbf{Update}
&
$ \boldsymbol{\psi}'=\boldsymbol{\psi}+\mathbf{b} $
&
$ \boldsymbol{\psi}'_c=Q_c\boldsymbol{\psi}_c $
&
$ \boldsymbol{\psi}'_c=Q_c\boldsymbol{\psi}_c $
\\

\textbf{Generator}
&
--
&
$ S_c=0.5(P_c-P_c^\top) $
&
$ S_c=\mathbf{u}_c\mathbf{v}_c^\top-\mathbf{v}_c\mathbf{u}_c^\top $
\\

\textbf{Cayley map}
&
--
&
$ Q_c=(I-S_c)^{-1}(I+S_c) $
&
$ Q_c=(I-S_c)^{-1}(I+S_c) $
\\

\textbf{Parameters}
&
$ nC $
&
$ Cn^2 $
&
$ 2nC $
\\

\textbf{Orthogonal}
&
No
&
Yes
&
Yes
\\

\textbf{Preserves norm}
&
No
&
Yes
&
Yes
\\

\textbf{Geometry}
&
Latent translation
&
Dense phase rotation
&
Rank-2 planar phase rotation
\\

\textbf{Effect}
&
Entangles phase and magnitude
&
Expressive but parameter-heavy
&
Parameter-efficient orthogonal update
\\

\bottomrule
\end{tabular}
}
\end{table}

\subsection{Rank-2 per-channel orthogonal updates}

A dense Cayley parameterization~\cite{yuan2024bridging} represents each phase update using a full skew-symmetric matrix \(S \in \mathbb{R}^{n \times n}\). While the Cayley transform guarantees that the resulting operator is orthogonal, this approach requires \(\mathcal{O}(n^2)\) parameters per rotation operator and an \(n \times n\) linear solve. This cost becomes increasingly undesirable as the size of the rotation dimension \(n\) grows. We therefore investigate a more parameter-efficient alternative based on low-rank skew-symmetric generators.

Let the rotating activation tensor after applying a neural layer be \(\boldsymbol{\psi} = f_{\mathbf w}(\mathbf z_{\mathrm{in}})  \in \mathbb{R}^{B \times n \times C \times H \times W}\), where \(n\) is the rotation dimension and \(C\) is the number of feature channels. CORE learns a separate low-rank orthogonal update for each channel. Specifically, we introduce two learnable tensors \(\mathbf{u}, \mathbf{v} \in \mathbb{R}^{C \times n}\), so that each channel \(c\) is associated with vectors \(\mathbf{u}_c,\mathbf{v}_c \in \mathbb{R}^{n}\). These vectors define a rank-2 skew-symmetric generator:
\begin{equation}
    S_c
    =
    \mathbf{u}_c \mathbf{v}_c^\top
    -
    \mathbf{v}_c \mathbf{u}_c^\top
    \in \mathbb{R}^{n \times n}.
\end{equation}
This construction is skew-symmetric by design, since \(S_c^\top = -S_c\). Applying the Cayley transform yields the channel-specific orthogonal matrix:
\begin{equation}
    Q_c
    =
    (I - S_c)^{-1}(I + S_c),
\end{equation}
which satisfies \(Q_c^\top Q_c = I\). Each feature channel therefore receives its own norm-preserving phase transformation, enabling channel-specific phase synchronization without relying on additive translations.
Overall, the difference between RF~\cite{lowe2023rotating} and CORE can be summarized as the choice of phase update applied to the layer response \(f_{\mathbf w}(\mathbf z_{\mathrm{in}})\) in Equation~\ref{eq:conv}:

\begin{equation}
\boldsymbol{\psi}
\;\xrightarrow[\text{CORE}]{\text{RF}}\;
\begin{cases}
f_{\mathbf w}(\mathbf z_{\mathrm{in}}) + \mathbf b, & \text{additive phase shift},\\[4pt]
Q\,f_{\mathbf w}(\mathbf z_{\mathrm{in}}), & \text{orthogonal phase transformation},
\end{cases}
\qquad \in \mathbb{R}^{n \times d_{\mathrm{out}}}.
\label{eq:rf_core_transition}
\end{equation}


The rank-2 form substantially reduces the number of learnable parameters. A full per-channel skew-symmetric matrix has \(n(n-1)/2\) independent degrees of freedom, requiring \(C n(n-1)/2\) parameters across all channels. In practice, dense Cayley implementations~\cite{yuan2024bridging} often learn an unconstrained matrix \(P_c \in \mathbb{R}^{n \times n}\) and form \(S_c = 0.5(P_c - P_c^\top)\), which uses \(Cn^2\) trainable parameters. CORE instead learns only two vectors of length \(n\) per channel, giving \(2nC\) parameters. Relative to the minimal full skew-symmetric parameterization, the parameter ratio is \((2nC)/(C n(n-1)/2) = 4/(n-1)\), while relative to the dense unconstrained implementation it is \((2nC)/(Cn^2) = 2/n\). Thus, CORE reduces the learnable parameter cost from \(\mathcal{O}(Cn^2)\) to \(\mathcal{O}(Cn)\) in practice, while retaining exact orthogonality through the Cayley transform. In the current implementation, the full \(n \times n\) matrix \(Q_c\) is still constructed and evaluated through a linear solve; therefore, CORE primarily improves parameter efficiency rather than the computational complexity of the Cayley transform itself. The main differences between the RF~\cite{lowe2023rotating} update, the dense Cayley formulation, and our rank-2 CORE implementation are summarized in Table~\ref{tab:phase_update_comparison}.

\section{Results}
\label{results}
\begin{table}[t]
\caption{
Object discovery results on the 4Shapes RGB dataset, reported as mean \(\pm\) standard deviation over 4 seeds. CORE outperforms RF and maintains consistent grouping performance across color-set variants. Best and second-best values are shown as \bestcap{bold} and \secondcap{underlined}, respectively.}
\label{tab:4shapes_rgb}
\centering
\renewcommand{\arraystretch}{1.2}
\begin{threeparttable}
\begin{tabular}{l c c c c c}
\hline
\textbf{Model} & \textbf{Dataset} & \textbf{Colors} & \textbf{MSE ↓} & \textbf{ARI-BG ↑} & \textbf{MBO\textsubscript{i} ↑} \\
\hline

\multirow{5}{*}{RF}
& RGB & 1 & \bestsci{1.017}{-5}{\sci{2.541}{-6}} & \secondvalerr{0.982}{0.009} & \secondvalerr{0.874}{0.008} \\
& RGB & 2 & \bestsci{1.597}{-4}{\sci{2.185}{-5}} & \secondvalerr{0.726}{0.035} & \secondvalerr{0.595}{0.015} \\
& RGB & 3 & \bestsci{1.044}{-4}{\sci{6.547}{-6}} & \secondvalerr{0.850}{0.008} & \secondvalerr{0.767}{0.016} \\
& RGB & 4 & \bestsci{1.166}{-4}{\sci{5.985}{-6}} & \secondvalerr{0.682}{0.041} & \secondvalerr{0.626}{0.050} \\
& RGB & 5 & \bestsci{1.653}{-4}{\sci{3.065}{-5}} & \secondvalerr{0.494}{0.056} & \secondvalerr{0.474}{0.031} \\
\hline

\multirow{5}{*}{CORE}
& RGB & 1 & \secondsci{2.500}{-4}{\sci{7.071}{-5}} & \bestvalerr{0.999}{0.000} & \bestvalerr{0.999}{0.000} \\
& RGB & 2 & \secondsci{8.000}{-4}{\sci{1.001}{-5}} & \bestvalerr{0.996}{0.000} & \bestvalerr{0.981}{0.001} \\
& RGB & 3 & \secondsci{1.100}{-3}{\sci{4.243}{-4}} & \bestvalerr{0.995}{0.002} & \bestvalerr{0.981}{0.005} \\
& RGB & 4 & \secondsci{2.050}{-3}{\sci{9.192}{-4}} & \bestvalerr{0.985}{0.011} & \bestvalerr{0.957}{0.023} \\
& RGB & 5 & \secondsci{1.700}{-3}{\sci{2.828}{-4}} & \bestvalerr{0.991}{0.001} & \bestvalerr{0.966}{0.008} \\
\hline

\end{tabular}
\end{threeparttable}
\end{table}

\begin{figure}[t]
    \centering
    \includegraphics[width=1.0\linewidth]{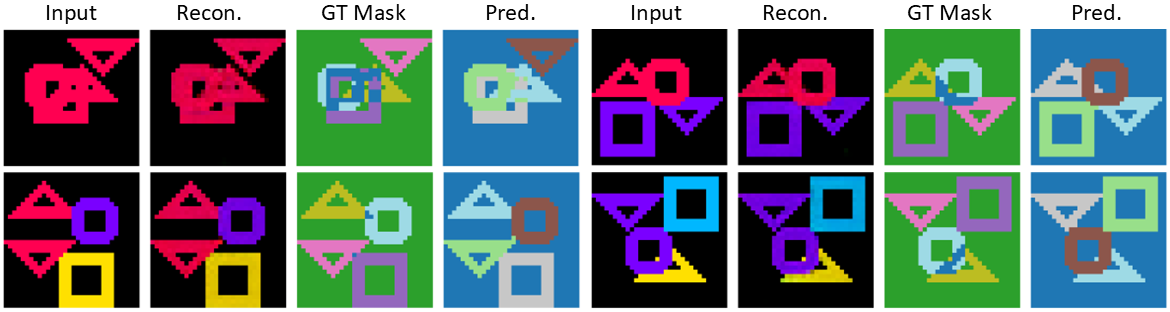}
    \caption{Qualitative results of CORE on the 4Shapes RGB dataset.}
    \label{fig:4shapes}
\end{figure}

In this section, we present our experimental results. We first describe the training procedure, the evaluation metrics and the datasets. We then quantitatively compare the grouping performance of CORE against state-of-the-art synchrony-based and object-centric baselines, including CAE~\cite{lowe2022complex}, CAE++~\cite{stanic2023contrastive}, CtCAE~\cite{stanic2023contrastive}, RF~\cite{lowe2023rotating}, SynCx~\cite{gopalakrishnan2024recurrent}, and Slot Attention~\cite{locatello2020object}, on unsupervised object discovery tasks. Next, we assess the contribution of key model components through ablation experiments. 

\paragraph{General setup}
We implement CORE as a convolutional autoencoding architecture following the RF~\cite{lowe2023rotating} setup. Apart from our proposed phase-update mechanism, the main architectural changes are the removal of the final sigmoid layer and the replacement of transposed convolutions in the decoder with nearest-neighbor upsampling followed by convolutional layers, similar to~\citet{stanic2023contrastive}.
Each model is trained with a batch size of 16 for 40,000--50,000 steps, depending on the dataset. We use the Adam optimizer~\cite{kingma2014adam} with a cosine annealing learning-rate schedule. All experiments are implemented in PyTorch~\cite{paszke2019pytorch} and run on a single NVIDIA H100 GPU. Additional details on the experimental setup are provided in the Appendix.

\begin{wraptable}{r}{0.48\textwidth}
\vspace{-1em}
\caption{
Object discovery results on the Tetrominoes dataset, reported as mean \(\pm\) standard deviation over 5 seeds. CORE outperforms the other synchrony-based models and achieves performance comparable to Slot Attention~\cite{locatello2020object}. Best and second-best values are shown as \bestcap{bold} and \secondcap{underlined}, respectively.}
\label{tab:tetr}
\centering
\renewcommand{\arraystretch}{1.2}
\begin{threeparttable}
\resizebox{0.48\textwidth}{!}{
\begin{tabular}{l c c}
\hline
\textbf{Model} & \textbf{ARI-BG ↑} & \textbf{MBO\textsubscript{i} ↑} \\
\hline
\multicolumn{3}{l}{\textit{Synchrony-based models}} \\

CAE~\cite{lowe2022complex}
& \valerr{0.000}{0.000}
& -- \\

CAE++~\cite{stanic2023contrastive}
& \valerr{0.780}{0.070}
& -- \\

CtCAE~\cite{stanic2023contrastive}
& \valerr{0.840}{0.090}
& -- \\

SynCx~\cite{gopalakrishnan2024recurrent}
& \valerr{0.890}{0.001}
& -- \\

RF~\cite{lowe2023rotating}
& \valerr{0.420}{0.060}
& \valerr{0.440}{0.080} \\

ItrSA~\cite{miyato2024artificial}
& \valerr{0.858}{0.008}
& \valerr{0.549}{0.003} \\

AKOrN~\cite{miyato2024artificial}
& \valerr{0.886}{0.002}
& \secondvalerr{0.564}{0.009} \\

AKOrN{\scriptsize\textit{+up-tiling}}~\cite{miyato2024artificial}
& \secondvalerr{0.931}{0.003}
& \valerr{0.563}{0.000} \\

CORE
& \bestvalerr{0.976}{0.001}
& \bestvalerr{0.976}{0.002} \\

\hline
\multicolumn{3}{l}{\textit{Slot-based model}} \\

Slot Attention~\cite{locatello2020object}
& \bestvalerr{0.990}{0.200}\textsuperscript{\(\dagger\)}
& -- \\

\hline
\end{tabular}
}
\begin{tablenotes}
\footnotesize
\item \(\dagger\) denotes a run excluded from evaluation as an outlier.
\end{tablenotes}
\end{threeparttable}
\vspace{-1em}
\end{wraptable}

\paragraph{Evaluation metrics}
We follow the evaluation protocol of prior work~\cite{greff2019multi,locatello2020object,emami2021efficient} for assessing grouping performance. Reconstruction quality is measured using MSE. To evaluate unsupervised object discovery, we report Adjusted Rand Index (ARI)~\cite{rand1971objective,hubert1985comparing} and Mean Best Overlap (MBO)~\cite{pont2016multiscale,seitzer2022bridging}. Following standard practice, we report ARI-BG over foreground pixels and ARI+BG over all pixels, including background. For MBO, we report two variants: instance-level MBO\textsubscript{i} and class-level MBO\textsubscript{c}.

\begin{figure}[t]
    \centering
    \includegraphics[width=1.0\linewidth]{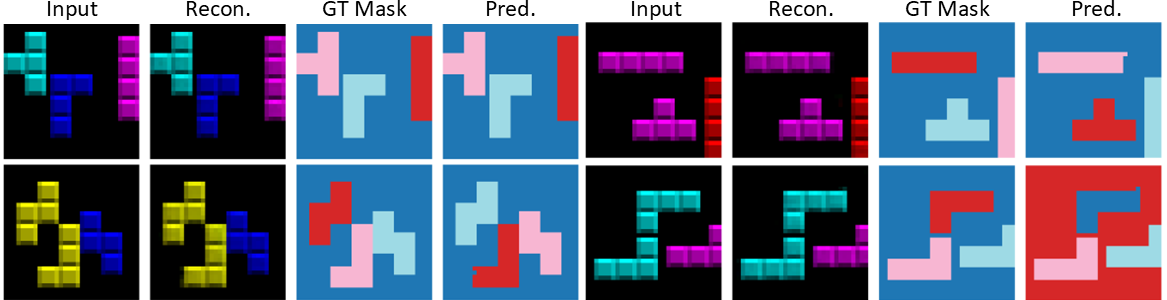}
    \caption{Qualitative results of CORE on the Tetrominoes dataset.}
    \label{fig:tetro}
\end{figure}

\paragraph{Datasets}
We evaluate the models on both synthetic and real-image datasets. The 4Shapes RGB(-D) dataset~\cite{lowe2023rotating} consists of \(32 \times 32\) synthetic images containing four distinct shapes: a square, an upward-pointing triangle, a downward-pointing triangle, and a circle, placed on a black background. To assess robustness to color diversity, we use five publicly available variants with increasing color-set size, ranging from a single shared color to five distinct colors. The dataset includes both RGB and RGB-D variants, where each object is assigned a fixed depth value. 
In our main experiments, CORE is trained and evaluated using only RGB inputs; depth information is not used. We also evaluate on Tetrominoes~\cite{multiobjectdatasets19}, which contains \(35 \times 35\) synthetic images with three tetromino-shaped objects. Following our experimental setup, we center-crop the images to \(32 \times 32\). Finally, we evaluate CORE on real images from Pascal VOC dataset~\cite{everingham2010pascal}. 

\newcommand{\bestval}[1]{\textbf{#1}}
\newcommand{\secondval}[1]{\underline{#1}}

\begin{wraptable}{r}{0.56\textwidth}
\vspace{-2em}
\caption{
Object discovery performance on Pascal VOC. Results are reported using MBO\textsubscript{i} and MBO\textsubscript{c}. 
RF and CORE are evaluated using pretrained DINO features as input. 
For RF and CORE, we report different architectural setups. 
Best and second-best values are shown as \bestcap{bold} and \secondcap{underlined}, respectively.
}
\label{tab:pascal_voc}
\centering
\renewcommand{\arraystretch}{1.15}
\begin{threeparttable}
\resizebox{0.56\textwidth}{!}{
\begin{tabular}{l c c c c}
\hline
\textbf{Model} & \textbf{Batch} & \textbf{Decoder} & \textbf{MBO\textsubscript{i} $\uparrow$} & \textbf{MBO\textsubscript{c} $\uparrow$} \\
\hline

\multicolumn{5}{l}{\textit{Slot-based models}} \\

Slot Attention~\cite{locatello2020object}
& -- & -- & 0.222 & 0.237 \\

SLATE~\cite{singh2021illiterate}
& -- & -- & 0.359 & 0.415 \\

\hline
\multicolumn{5}{l}{\textit{DINO-feature models}} \\

DINO $k$-means
& -- & -- & 0.363 & 0.405 \\

DINOSAUR~\cite{seitzer2022bridging}
& 64 & -- & 0.440 & 0.512 \\

DINOSAUR MLP~\cite{seitzer2022bridging}
& 64 & -- & 0.395 & 0.409 \\

Slot Diffusion~\cite{wu2023slotdiffusion}
& 64 & -- & \secondval{0.504} & \secondval{0.553} \\

SPOT~\cite{kakogeorgiou2024spot}
& 64 & -- & 0.483 & 0.556 \\

\hline
\multicolumn{5}{l}{\textit{DINO-feature synchrony models}} \\

DINO CAE~\cite{lowe2022complex}
& 64 & transpose & 0.329 & 0.374 \\

RF\textsuperscript{\(\dagger\)}~\cite{lowe2023rotating}
& 64 & transpose
& 0.407
& 0.460 \\

RF~\cite{lowe2023rotating}
& 16 & transpose
& 0.394
& 0.443 \\

RF~\cite{lowe2023rotating}
& 16 & upsampling
& 0.293
& 0.338 \\

CORE
& 16 & transpose
& 0.382
& 0.435 \\

CORE\textsuperscript{\(\dagger\)}
& 16 & upsampling
& 0.408
& 0.462 \\

\hline
\multicolumn{5}{l}{\textit{Synchrony-based model}} \\
AKOrN~\cite{miyato2024artificial}
& 512 & --
& \bestval{0.520}
& \bestval{0.603} \\

\hline
\end{tabular}
}
\begin{tablenotes}
\footnotesize
\item \(\dagger\) indicates the default configuration for each model.
\end{tablenotes}
\end{threeparttable}
\vspace{-2em}
\end{wraptable}

\paragraph{Unsupervised object discovery in multi-channel images}
We evaluate the applicability of CORE to multi-channel inputs using the 4Shapes RGB(-D) dataset. As shown in Table~\ref{tab:4shapes_rgb}, RF struggles on the RGB version of 4Shapes as the number of possible object colors increases. This behavior is consistent with the observation in the RF paper~\cite{lowe2023rotating} that the model can incorrectly group different shapes based on shared color rather than object identity. In the Appendix, we report RF results using depth masks, where this ambiguity is largely resolved and RF successfully separates the objects. In contrast, CORE does not require depth information and maintains consistently strong grouping performance across color-set variants. Qualitative results in Figure~\ref{fig:4shapes} further support the quantitative findings.

We also evaluate CORE on the Tetrominoes dataset against other synchrony-based models and Slot Attention, as shown in Table~\ref{tab:tetr} and Figure~\ref{fig:tetro}. Among the synchrony-based baselines, CAE~\cite{lowe2022complex}, CAE++~\cite{stanic2023contrastive}, CtCAE~\cite{stanic2023contrastive}, and RF~\cite{lowe2023rotating} use a binding mechanism~\cite{reichert2013neuronal} closely related to CORE. In contrast, SynCx~\cite{gopalakrishnan2024recurrent} uses a recurrent complex-valued weight model, ItrSA~\cite{miyato2024artificial} relies on iterative self-attention layers, and AKOrN~\cite{miyato2024artificial} is inspired by Kuramoto oscillator dynamics. The up-tiling variant of AKOrN applies an upsampling procedure to obtain finer cluster assignments, which substantially improves its object discovery performance.

\begin{wraptable}{r}{0.56\textwidth}
\vspace{-1em}
\caption{
Ablation study on the effect of sigmoid and upsampling on the Tetrominoes dataset. 
Removing the sigmoid substantially improves performance for both RF and CORE, while upsampling further improves object discovery, especially for CORE. 
Best and second-best values are shown as \bestcap{bold} and \secondcap{underlined}, respectively. 
}
\label{tab:ablation_sigmoid_upsample}
\centering
\renewcommand{\arraystretch}{1.15}
\begin{threeparttable}
\resizebox{0.56\textwidth}{!}{
\begin{tabular}{l c c c c c}
\hline
\textbf{Model} & \textbf{Sigmoid} & \textbf{Upsample} & \textbf{ARI+BG $\uparrow$} & \textbf{ARI-BG $\uparrow$} & \textbf{MBO\textsubscript{i} $\uparrow$} \\
\hline

\multirow{4}{*}{RF}
& yes\textsuperscript{\(\dagger\)} & no\textsuperscript{\(\dagger\)}
& 0.657
& 0.127
& 0.369 \\

& yes & yes
& 0.566
& 0.394
& 0.490 \\

& no & no
& 0.974
& 0.762
& 0.817 \\

& no & yes
& \bestval{0.992}
& \secondval{0.923}
& \secondval{0.939} \\

\hline

\multirow{4}{*}{CORE}
& yes & no
& 0.441
& 0.036
& 0.281 \\

& yes & yes
& 0.537
& 0.017
& 0.294 \\

& no & no
& 0.950
& 0.365
& 0.533 \\

& no\textsuperscript{\(\dagger\)} & yes\textsuperscript{\(\dagger\)}
& \secondval{0.992}
& \bestval{0.977}
& \bestval{0.978} \\

\hline
\end{tabular}
}
\begin{tablenotes}
\footnotesize
\item \(\dagger\) indicates the default configuration for each model.
\end{tablenotes}
\end{threeparttable}
\vspace{-1em}
\end{wraptable}
Among all synchrony-based models, CORE achieves the best performance, with an ARI-BG of 0.976 and an MBO\textsubscript{i} of 0.976. This result is close to Slot Attention~\cite{locatello2020object} and establishes a new state-of-the-art among synchrony-based models on Tetrominoes.

\begin{wraptable}{r}{0.56\textwidth}
\vspace{-1em}
\caption{
Ablation study on the effect of sigmoid and upsampling on the Tetrominoes dataset. 
Removing the sigmoid substantially improves performance for both RF and CORE, while upsampling further improves object discovery, especially for CORE. 
Best and second-best values are shown as \bestcap{bold} and \secondcap{underlined}, respectively. 
}
\label{tab:ablation_sigmoid_upsample}
\centering
\renewcommand{\arraystretch}{1.15}
\begin{threeparttable}
\resizebox{0.56\textwidth}{!}{
\begin{tabular}{l c c c c c}
\hline
\textbf{Model} & \textbf{Sigmoid} & \textbf{Upsample} & \textbf{ARI+BG $\uparrow$} & \textbf{ARI-BG $\uparrow$} & \textbf{MBO\textsubscript{i} $\uparrow$} \\
\hline

\multirow{4}{*}{RF}
& yes\textsuperscript{\(\dagger\)} & no\textsuperscript{\(\dagger\)}
& 0.657
& 0.127
& 0.369 \\

& yes & yes
& 0.566
& 0.394
& 0.490 \\

& no & no
& 0.974
& 0.762
& 0.817 \\

& no & yes
& \bestval{0.992}
& \secondval{0.923}
& \secondval{0.939} \\

\hline

\multirow{4}{*}{CORE}
& yes & no
& 0.441
& 0.036
& 0.281 \\

& yes & yes
& 0.537
& 0.017
& 0.294 \\

& no & no
& 0.950
& 0.365
& 0.533 \\

& no\textsuperscript{\(\dagger\)} & yes\textsuperscript{\(\dagger\)}
& \secondval{0.992}
& \bestval{0.977}
& \bestval{0.978} \\

\hline
\end{tabular}
}
\begin{tablenotes}
\footnotesize
\item \(\dagger\) indicates the default configuration for each model.
\end{tablenotes}
\end{threeparttable}
\vspace{-1em}
\end{wraptable}
\paragraph{Unsupervised object discovery in real images}
Table~\ref{tab:pascal_voc} and Figure~\ref{fig:pascal} report object discovery performance on the Pascal VOC dataset~\cite{everingham2010pascal}. Following the RF~\cite{lowe2023rotating} setup, we use a pretrained DINO model~\cite{caron2021emerging} to extract high-level input features and apply CORE directly to these representations. Under comparable batch-size settings, using the default architecture for each model—transposed convolutions for RF and upsampling for CORE—CORE slightly outperforms RF. Specifically, CORE achieves an MBO\textsubscript{i} of 0.408 compared to 0.394 for RF, and an MBO\textsubscript{c} of 0.462 compared to 0.443. AKOrN~\cite{miyato2024artificial}, another synchrony-based model, achieves the strongest overall performance; however, it uses a substantially larger batch size of 512, which is not feasible in our setting.

\paragraph{Ablation on architectural modifications}
Table~\ref{tab:ablation_sigmoid_upsample} presents an ablation study of the proposed architectural modifications on Tetrominoes. We find that applying a sigmoid activation at the decoder output substantially degrades performance on color datasets. We also observe a clear improvement when replacing transposed convolution layers with convolution followed by upsampling, especially for CORE. The final CORE architecture combines these modifications with our proposed orthogonal transformation mechanism. Together, these design choices yield substantially higher ARI and MBO\textsubscript{i} scores, supporting the effectiveness of the proposed architecture.

\paragraph{Parameter and training efficiency}

Table~\ref{tab:efficiency} compares the parameter count and training time of RF and CORE across datasets. CORE introduces only a modest increase in parameters relative to RF, due to the additional parameters required by the proposed orthogonal transformation mechanism, as discussed in Table~\ref{tab:phase_update_comparison}. For example, on 4Shapes, the parameter count increases by only 18,450 parameters, corresponding to about 1.44\%. CORE also has higher training cost due to the extra rotation operations. However, the increase remains moderate across datasets: even on the largest dataset, Pascal VOC, CORE requires only about 16 additional minutes of training compared to RF.




\section{Related Work}
\label{related_work}

\paragraph{Object-centric learning}
OCL~\cite{greff2020binding} aims to decompose visual scenes into structured object-level representations. A dominant class of methods~\cite{le2011learning,burgess2019monet,greff2019multi,eslami2016attend,kosiorek2018sequential,greff2016tagger,wu2023inverted,greff2019multi} uses slot-based representations, where a fixed set of latent slots provides an inductive bias for object discovery. Slot Attention~\citep{locatello2020object} uses iterative attention, allowing slots to compete for explaining different parts of the input; extensions such as SLATE~\citep{singh2021illiterate} and DINOSAUR~\citep{seitzer2022bridging} incorporate Transformer-based components~\citep{vaswani2017attention} to improve scalability and performance on more complex scenes. A complementary line of work studies object binding through neural synchrony: early synchronization-based grouping models~\citep{reichert2013neuronal} motivated deep architectures based on complex-valued activations, such as CAE~\citep{lowe2022complex}, and vector-valued rotating representations, such as RF~\citep{lowe2023rotating}. Recent approaches further explore biologically inspired spiking mechanisms~\citep{zheng2023gust}, contrastive synchronization objectives~\citep{stanic2023contrastive}, explicit phase synchronizers such as Kuramoto oscillators~\citep{miyato2024artificial,muzellec2025enhancing}, and recurrent architectures with complex-valued weights~\citep{gopalakrishnan2024recurrent}.

\begin{table}[t]
\caption{
Parameter count and training-time comparison between RF and CORE. Wall-clock training time is reported in minutes on a single NVIDIA H100 GPU.
}
\label{tab:efficiency}
\centering
\renewcommand{\arraystretch}{1.2}
\begin{tabular}{l c c c c c c}
\hline
\textbf{Model} 
& \multicolumn{3}{c}{\textbf{Parameters}} 
& \multicolumn{3}{c}{\textbf{Training Time (min)}} \\
\cline{2-4} \cline{5-7}
& \textbf{4Shapes} & \textbf{Tetrominoes} & \textbf{Pascal VOC}
& \textbf{4Shapes} & \textbf{Tetrominoes} & \textbf{Pascal VOC} \\
\hline

RF
& 1,284,660
& 1,543,711
& 12,240,846
& 11
& 13
& 40 \\

CORE
& 1,303,110
& 1,559,406
& 12,288,718
& 22
& 25
& 56 \\

\hline
\end{tabular}
\end{table}

\paragraph{Orthogonal transformations} 
Orthogonal transformations~\citep{qiu2023controlling,liu2023parameter} have been widely used for parameter-efficient fine-tuning. OFT~\citep{qiu2023controlling} learns orthogonal matrices, often parameterized with the Cayley transform~\citep{cayley1846quelques}, to preserve pairwise angles between neuron representations and limit deviations from the pretrained model. BOFT~\citep{liu2023parameter} improves efficiency using butterfly factorization, while qGOFT~\citep{ma2024parameter} reduces parameter cost further through quasi-Givens rotations~\citep{press2007numerical} and relaxed orthogonality constraints. Inspired by these methods, CORE applies orthogonal parameterization to rotating-feature models rather than weight fine-tuning: it learns rank-2 Cayley rotations along the rotation dimension, providing per-channel phase updates that preserve feature magnitude.

\section{Conclusion}
\label{conclusion}
\paragraph{Summary} In this work, we introduced CORE, a geometry-preserving extension of RF for unsupervised object discovery. CORE replaces additive latent biases with learned orthogonal transformations along the rotation dimension. Using a rank-2 Cayley parameterization, it updates object-affiliation signals through rotations while preserving feature magnitude. Experiments on synthetic RGB datasets and real images with DINO features show that CORE improves synchrony-based object discovery and supports the usefulness of structured phase updates.

\paragraph{Limitations and future work} CORE has some limitations. It introduces an increase in parameters compared to RF and requires longer training due to the additional rotation operations (see Table~\ref{tab:efficiency}). A promising direction for reducing this overhead is to exploit the low-rank structure of the learned transformations more directly, avoiding the explicit construction and inversion of full rotation matrices. This could preserve the same orthogonal update while reducing memory usage and training time. More broadly, future work could explore richer low-rank parameterizations and extend CORE to larger-scale real-world object discovery settings.

\begin{ack}
Use unnumbered first level headings for the acknowledgments. All acknowledgments
go at the end of the paper before the list of references. Moreover, you are required to declare
funding (financial activities supporting the submitted work) and competing interests (related financial activities outside the submitted work).
More information about this disclosure can be found at: \url{https://neurips.cc/Conferences/2026/PaperInformation/FundingDisclosure}.

Do {\bf not} include this section in the anonymized submission, only in the final paper. You can use the \texttt{ack} environment provided in the style file to automatically hide this section in the anonymized submission.
\end{ack}


\bibliographystyle{plainnat}
\bibliography{ref}

\medskip


\appendix

\section{Technical appendices and supplementary material}

\subsection{Broader Impact}
This work studies continuous and distributed object-centric representations and investigates how they can be extended from simple synthetic settings to real-world visual data. Its potential positive societal impacts include advancing methods for unsupervised object discovery, which may benefit areas such as computer vision, robotics, and representation learning. By building on biologically inspired ideas related to synchrony and binding, the work may also contribute to a better computational understanding of how object representations can emerge. In addition, the proposed structured representations may improve model interpretability and support the development of object-centric systems that generalize to more complex RGB images.

We do not anticipate direct negative societal impacts from this work. However, object discovery methods can have broad downstream applications, including in real-world vision systems. As a result, there may be unforeseen uses or consequences that are difficult to predict at this stage. We therefore acknowledge the possibility of indirect societal impacts and encourage careful consideration when applying such methods outside controlled research settings.

\subsection{Reproducibility Statement}
To ensure reproducibility, we provide a detailed description of the model architectures, hyperparameters, evaluation procedures, datasets, and baselines in Section~\ref{results} and the Appendix. We also plan to release the code upon acceptance. To obtain stable and reliable estimates, we run all experiments with 4--5 random seeds. Training and evaluation are performed on a single NVIDIA H100 GPU.

\subsection{Implementation Details}

We implement CORE within an autoencoding framework. In this section, we describe the architecture and hyperparameters used to train the model. The autoencoder architecture used across our experiments is detailed in Table~\ref{tab:autoencoder_architecture}, and the dataset-specific hyperparameters for all models are reported in Table~\ref{tab:hyperparameters}.

\begin{table*}[ht]
    \centering.
    \caption{CORE architecture with input dimensions \(h, w, c\) and feature dimension \(d\). All fractions are rounded up. We followed the same architecture as in RF \cite{lowe2023rotating}.}
    \label{tab:autoencoder_architecture}
    \renewcommand{\arraystretch}{1.2}
    \begin{tabular}{l c c c c c}
        \hline
         & \textbf{Layer} & \textbf{Feature Dimension} & \textbf{Kernel} & \textbf{Stride} & \textbf{Padding} \\ 
        & & & & \textbf{Input / Output} \\
        \hline
        & Input & \(h \times w \times c\) & - & - & - \\
        \hline
        \multirow{7}{*}{\textbf{Encoder}} 
        & Conv & \(h/2 \times w/2 \times d\) & 3 & 2 & 1 / 0 \\
        & Conv & \(h/2 \times w/2 \times d\) & 3 & 1 & 1 / 0 \\
        & Conv & \(h/4 \times w/4 \times 2d\) & 3 & 2 & 1 / 0 \\
        & Conv & \(h/4 \times w/4 \times 2d\) & 3 & 1 & 1 / 0 \\
        & Conv & \(h/8 \times w/8 \times 2d\) & 3 & 2 & 1 / 0 \\
        & Reshape & \(1 \times 1 \times (h/8 * w/8 * 2d)\) & - & - & - \\
        & Linear & \(1 \times 1 \times 2d\) & - & - & - \\
        \hline
        \multirow{7}{*}{\textbf{Decoder}}
        & Linear & \(1 \times 1 \times (h/8 * w/8 * 2d)\) & - & - & - \\
        & Reshape & \(h/8 \times w/8 \times 2d\) & - & - & - \\
        & TransConv & \(h/4 \times w/4 \times 2d\) & 3 & 2 & 1 / 1 \\
        & Conv & \(h/4 \times w/4 \times 2d\) & 3 & 1 & 1 / 0 \\
        & TransConv & \(h/2 \times w/2 \times d\) & 3 & 2 & 1 / 1 \\
        & Conv & \(h/2 \times w/2 \times d\) & 3 & 1 & 1 / 0 \\
        & TransConv & \(h \times w \times d\) & 3 & 2 & 1 / 1 \\
        \hline
    \end{tabular}
\end{table*}

\begin{table*}[ht]
    \centering
    \caption{Hyperparameters of the CORE for the three different datasets.}
    \label{tab:hyperparameters}
    \renewcommand{\arraystretch}{1.2}
    \begin{tabular}{l c c c}
        \hline
        \textbf{Dataset} & \textbf{Pascal VOC} & \textbf{Tetrominoes} & \textbf{4Shapes} \\
        \hline
        Training Steps & 30k & 60k & 100k \\
        Batch Size & 16 & 16 & 16 \\
        LR Warmup Steps & 1000 & 1000 & 1000  \\
        Peak LR & 0.001 & 0.0005 & 0.001 \\
        Gradient Norm Clipping & 0.1 & 0.1 & 0.1 \\
        \hline
        Feature dim \(d\) & 128 & 128 & 64 \\
        Rotating dim \(n\) & 10 & 4 & 5 \\
        Kernel size \(n\) & 3 & 5 & 5 \\
        \hline
        Image Size & 256 & 35 & 32 \\
        Crop Size &  224 & 32 & 32 \\
        Cropping Strategy & CenterCrop & CenterCrop & -  \\
        \hline
        Objects/Image & - & 3 & 4 \\
        No. Train Images & 45k & 40k & 50k  \\
        No. Test Images &  1449 & 320 & 5000  \\
        \hline
    \end{tabular}
\end{table*}

\subsection{Additional Results}
\begin{figure}
\setlength\tabcolsep{2pt}
\centering
\begin{tabular}{cc}
 \includegraphics[width=2.5in]{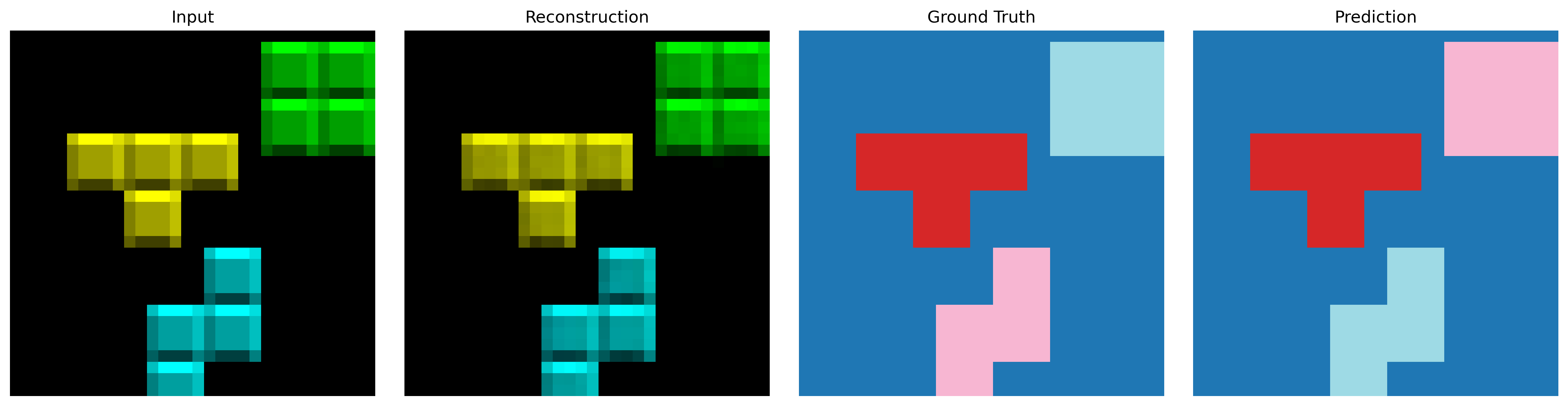} &
 \includegraphics[width=2.5in]{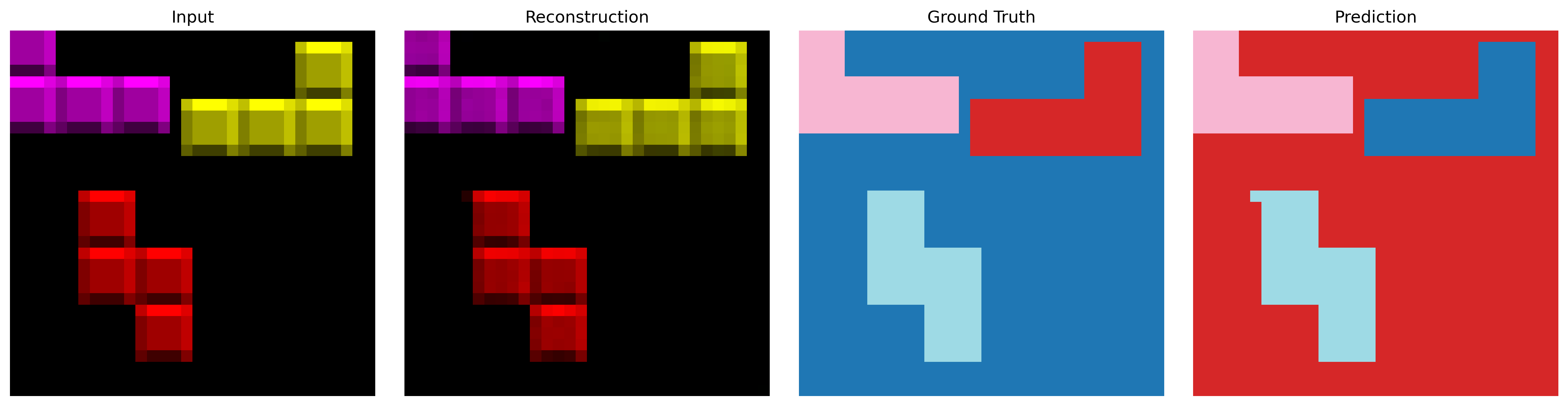} \\
 \includegraphics[width=2.5in]{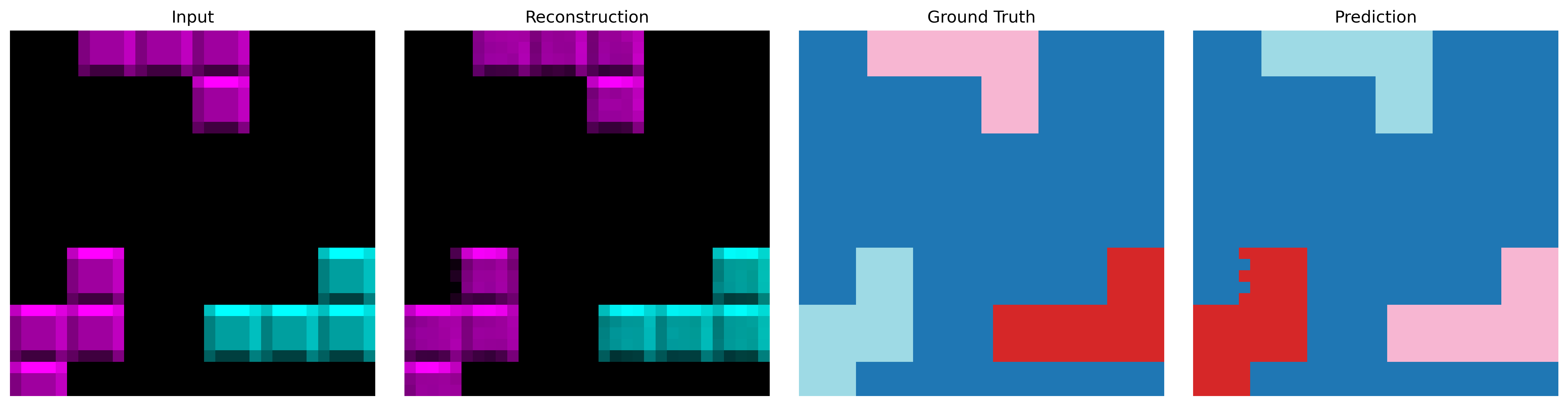} &
 \includegraphics[width=2.5in]{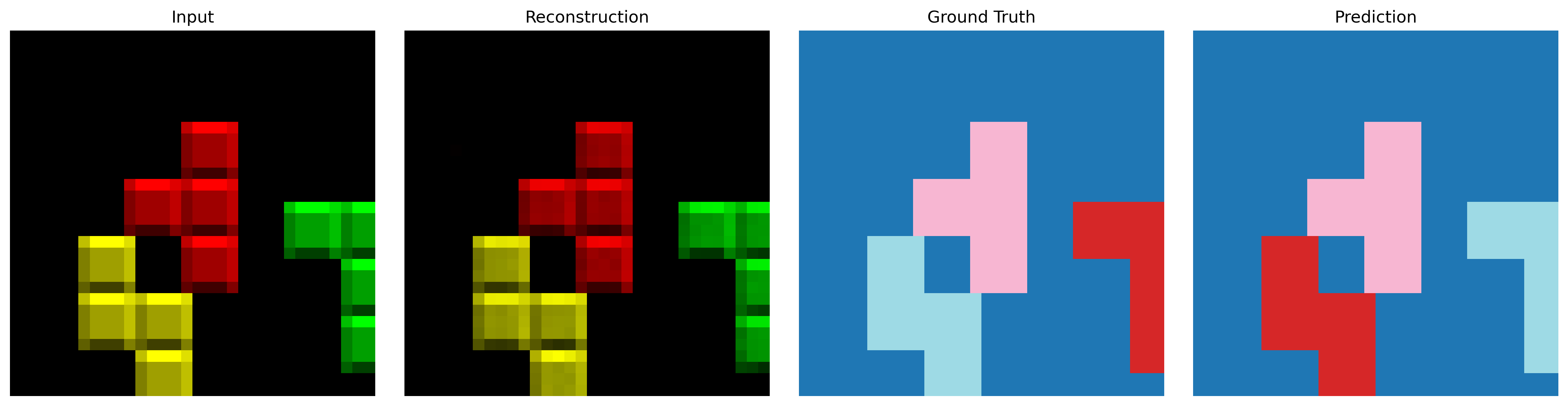} \\
 \includegraphics[width=2.5in]{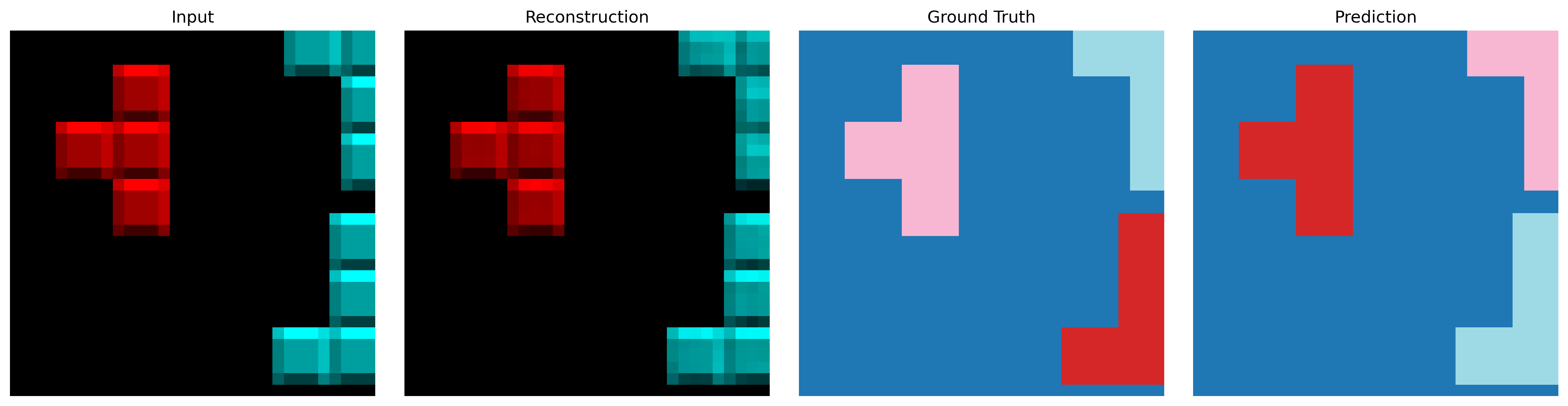} &
 \includegraphics[width=2.5in]{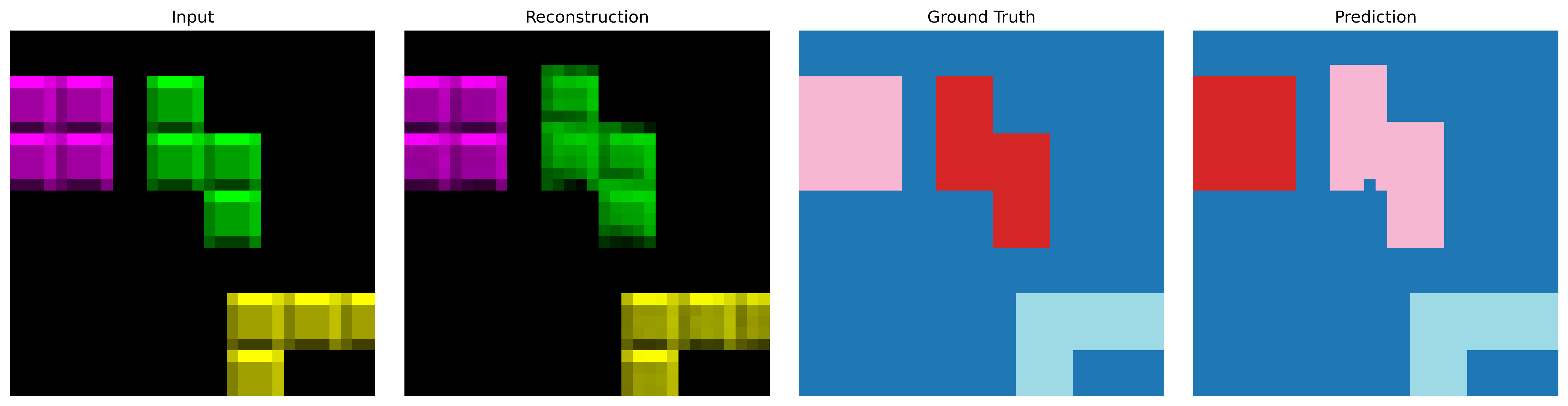} \\
  \includegraphics[width=2.5in]{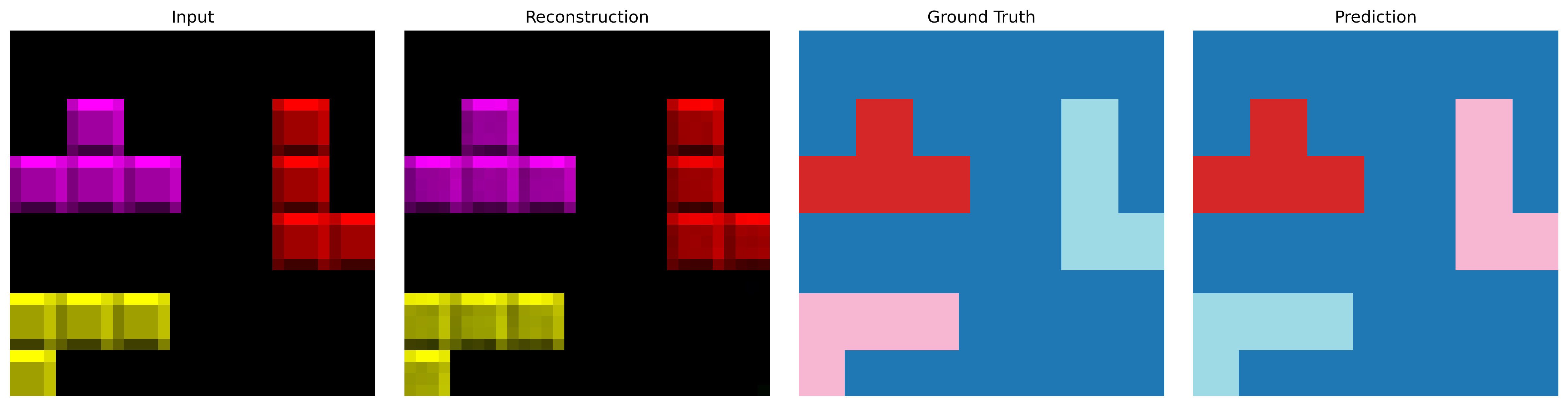} &
 \includegraphics[width=2.5in]{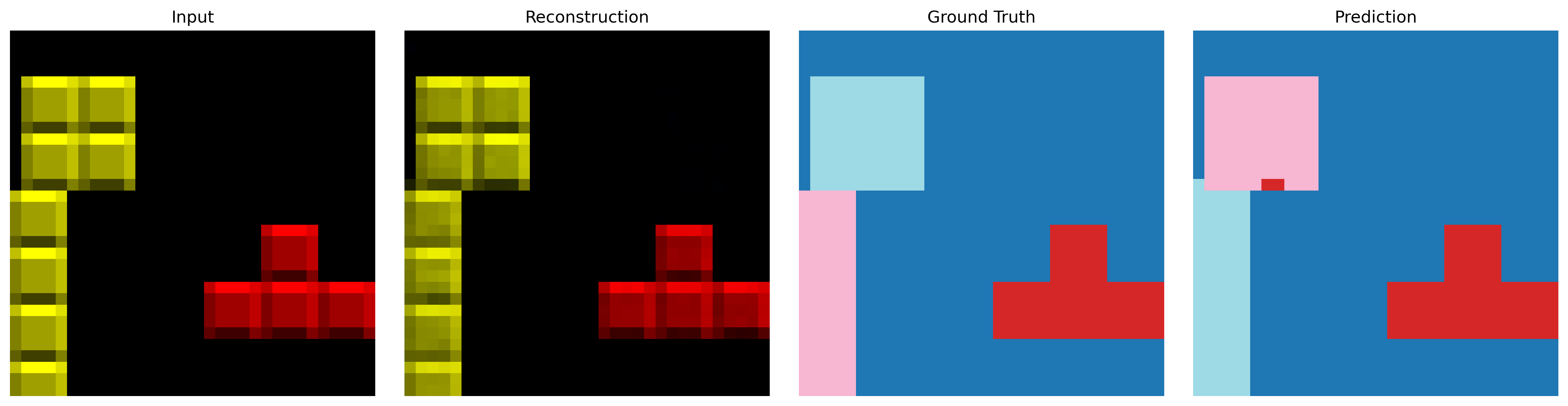} \\
 \includegraphics[width=2.5in]{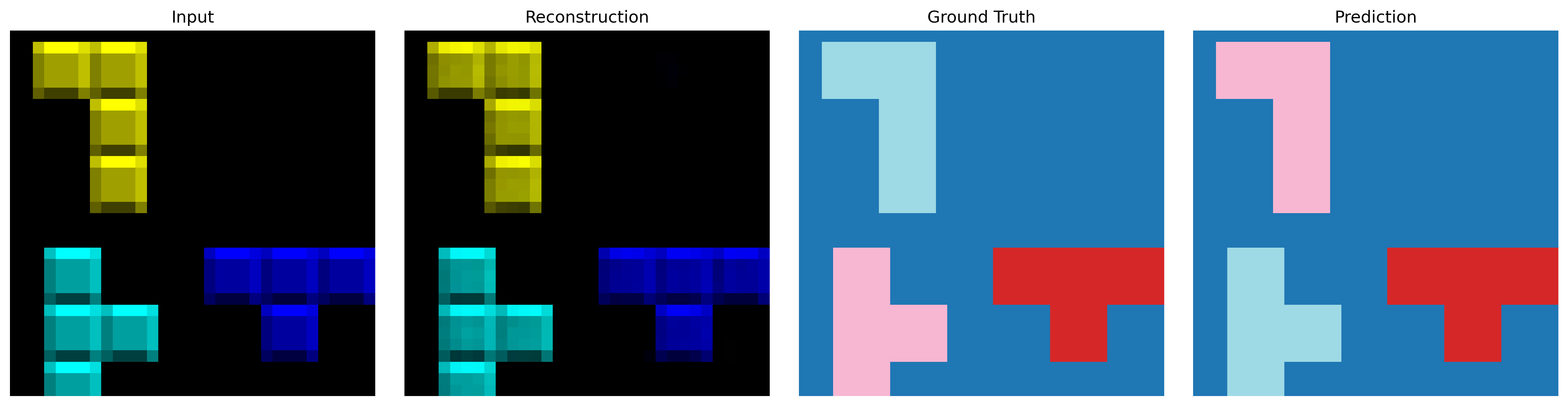} &
 \includegraphics[width=2.5in]{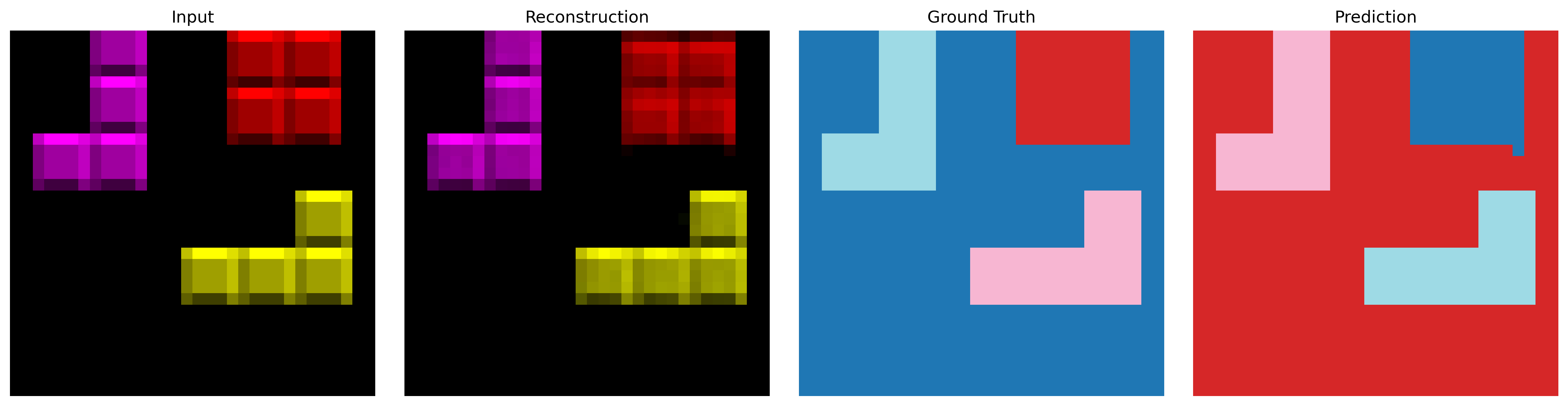} \\
 \includegraphics[width=2.5in]{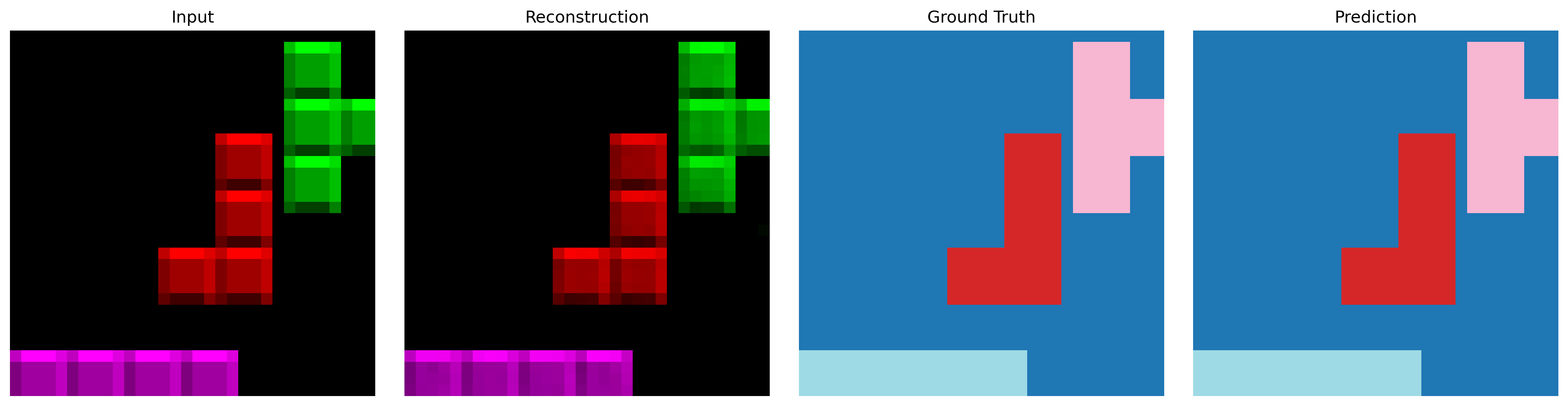} &
 \includegraphics[width=2.5in]{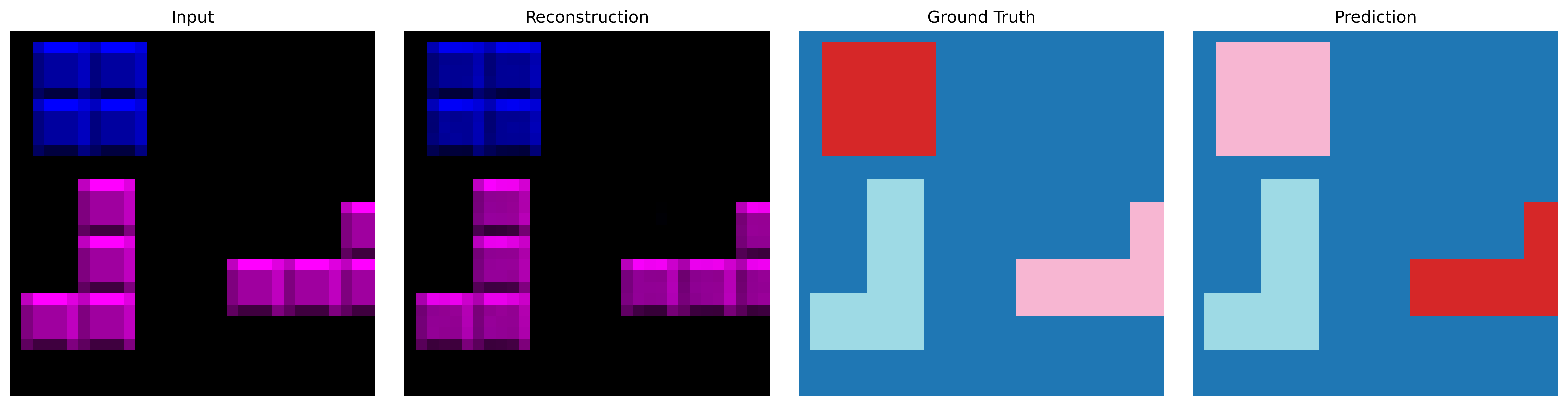} \\
  \includegraphics[width=2.5in]{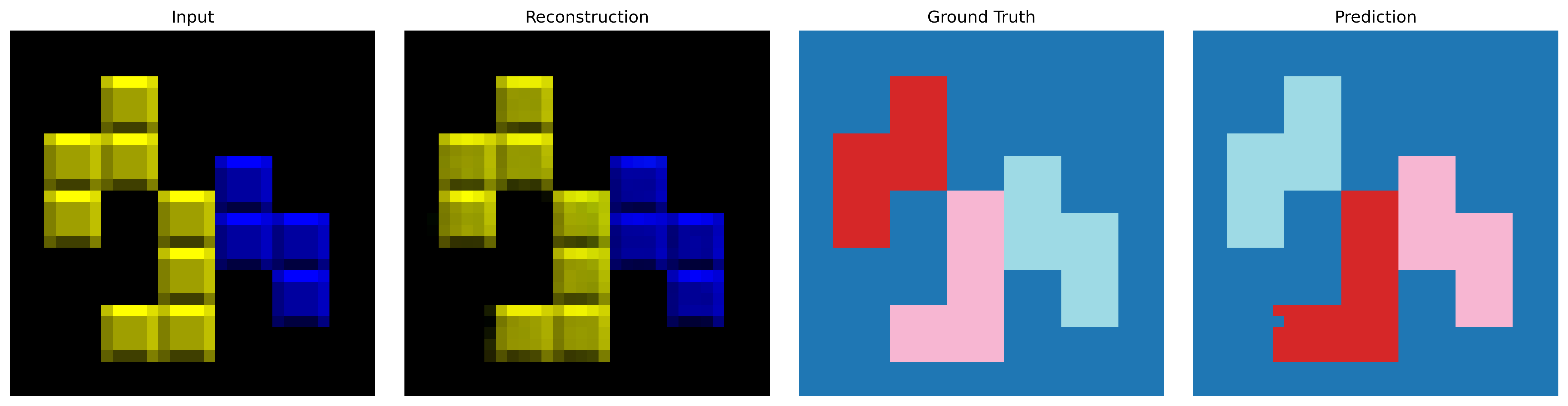} &
 \includegraphics[width=2.5in]{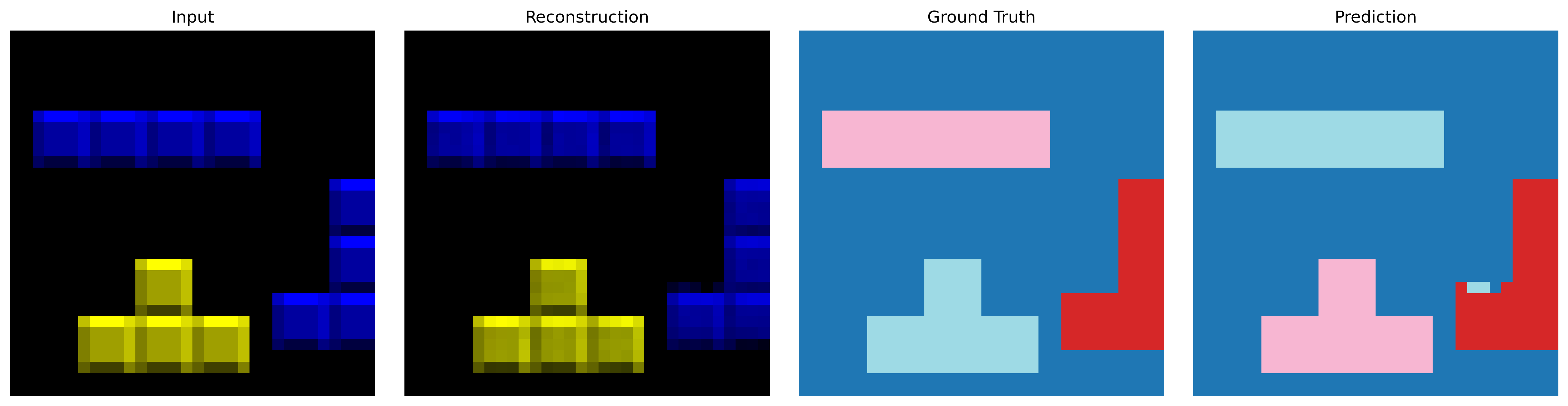} \\
 \includegraphics[width=2.5in]{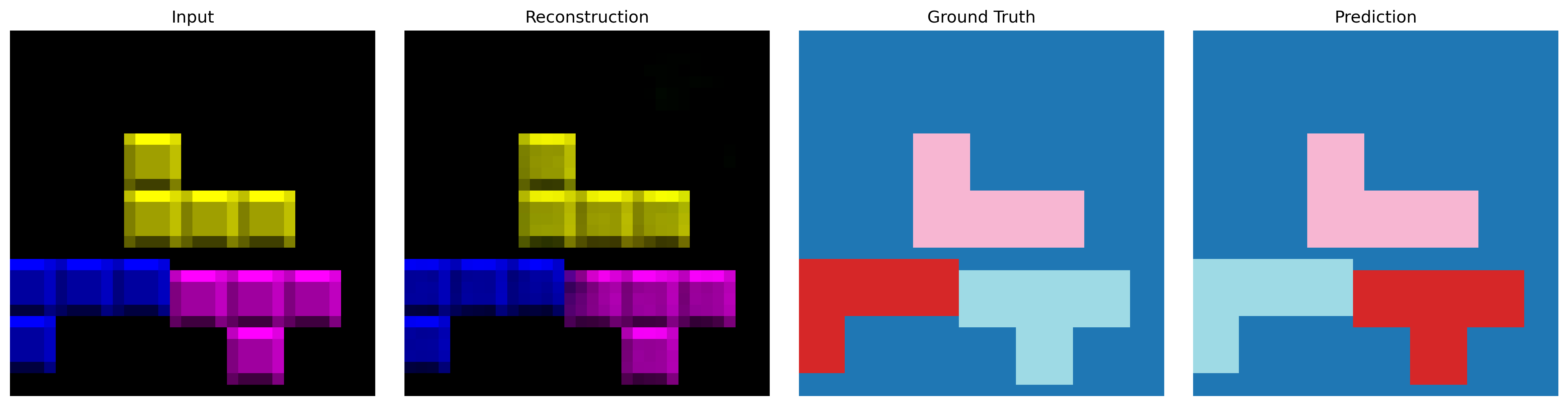} &
 \includegraphics[width=2.5in]{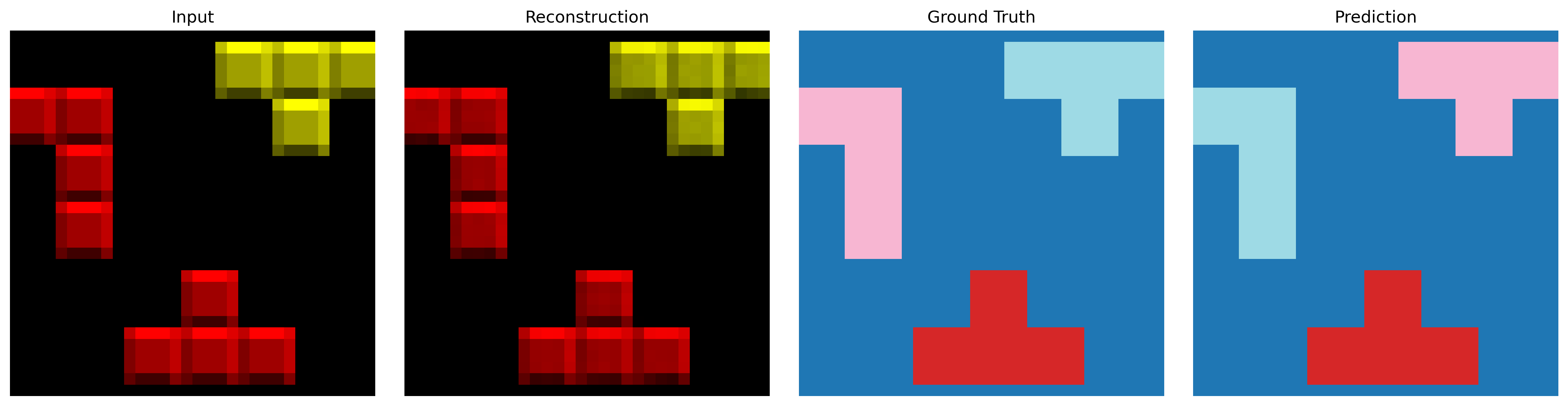} \\
 \includegraphics[width=2.5in]{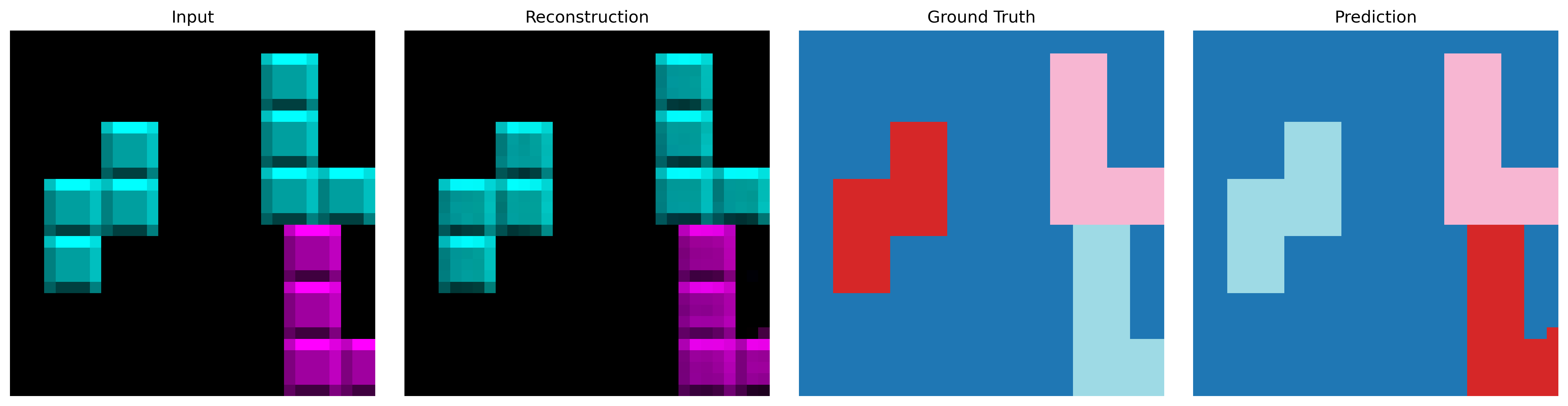} &
 \includegraphics[width=2.5in]{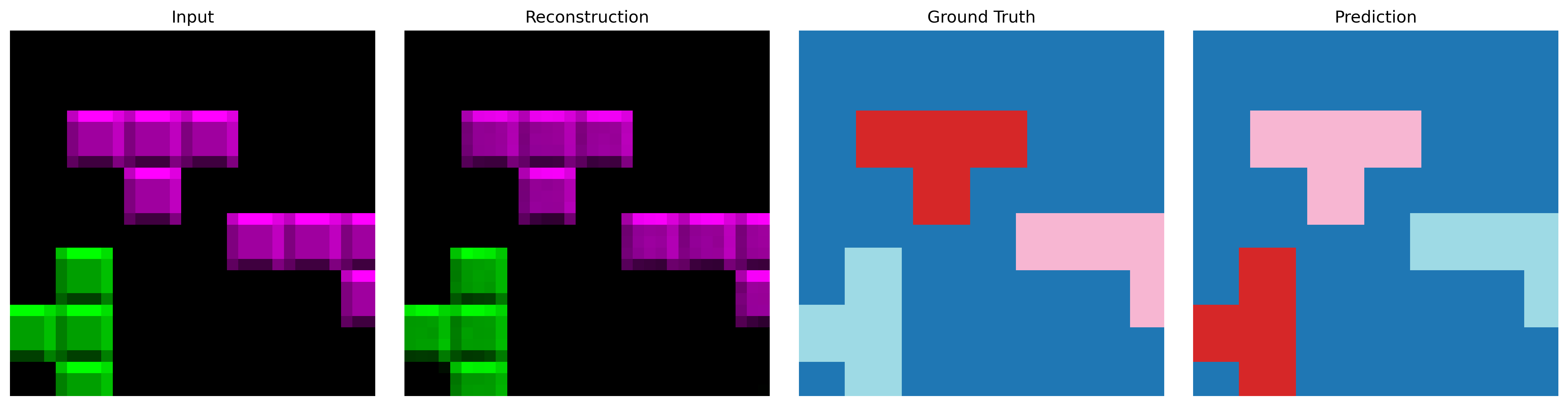} \\
  \includegraphics[width=2.5in]{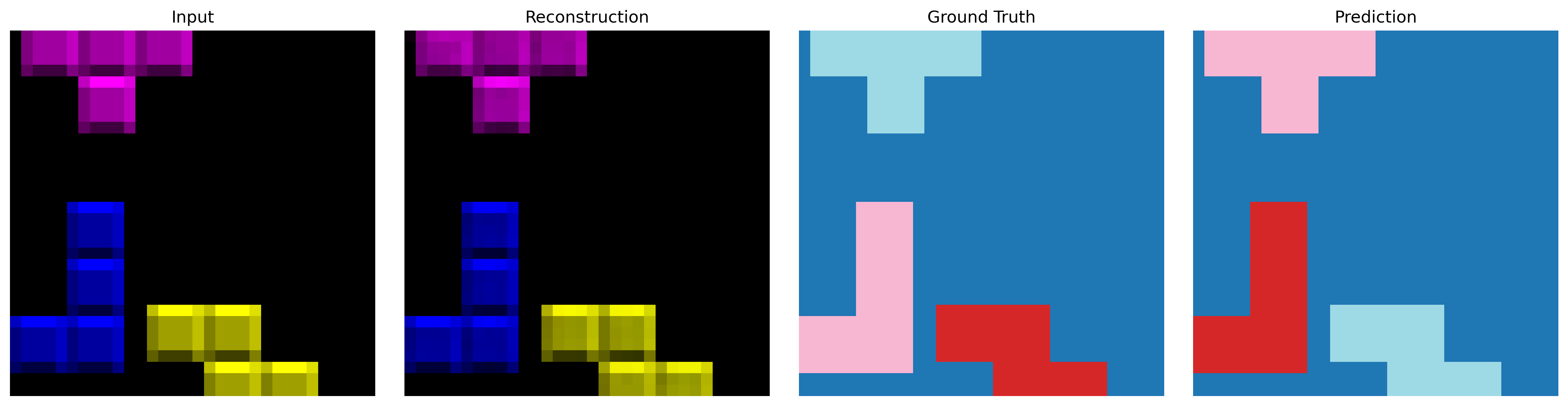} &
 \includegraphics[width=2.5in]{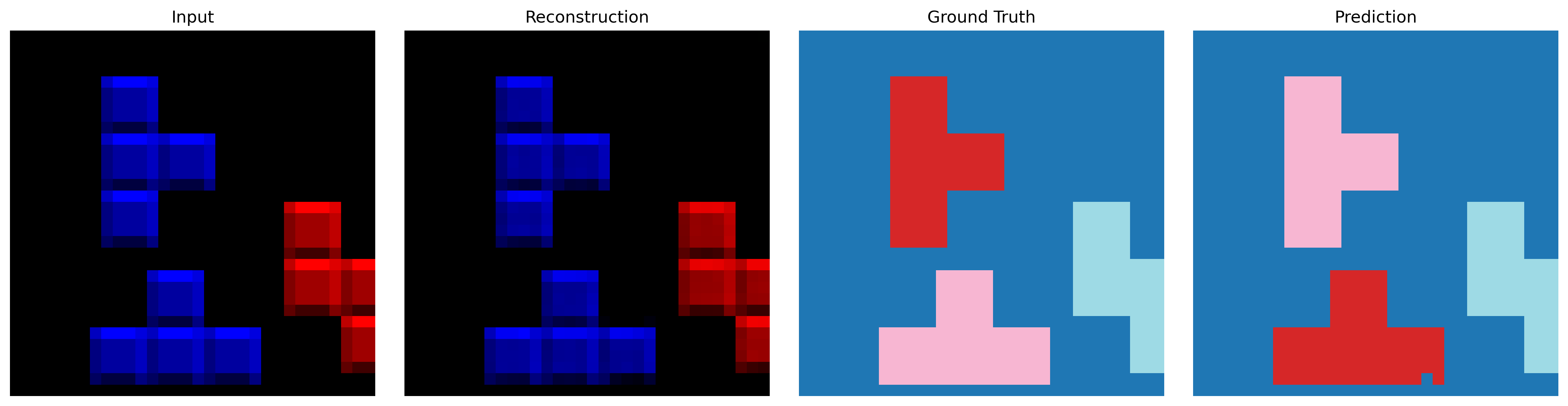} \\
 \includegraphics[width=2.5in]{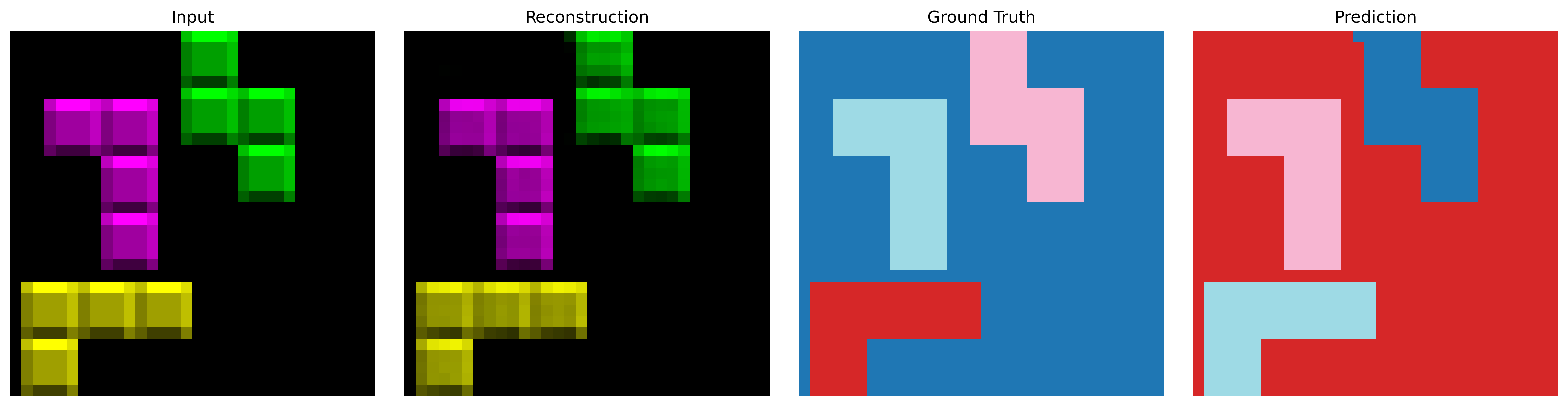} &
 \includegraphics[width=2.5in]{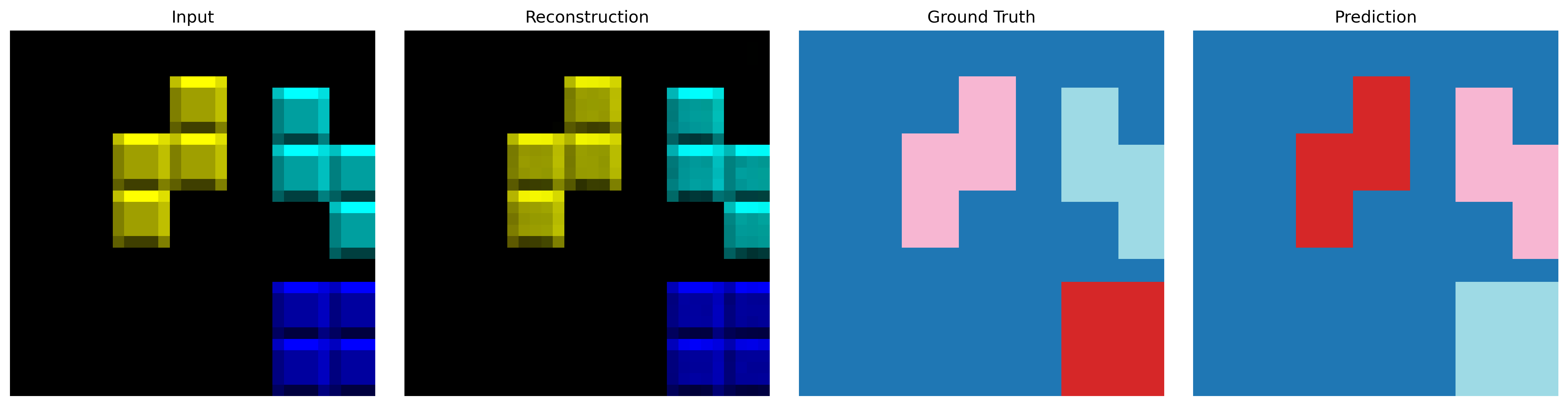} \\
 \includegraphics[width=2.5in]{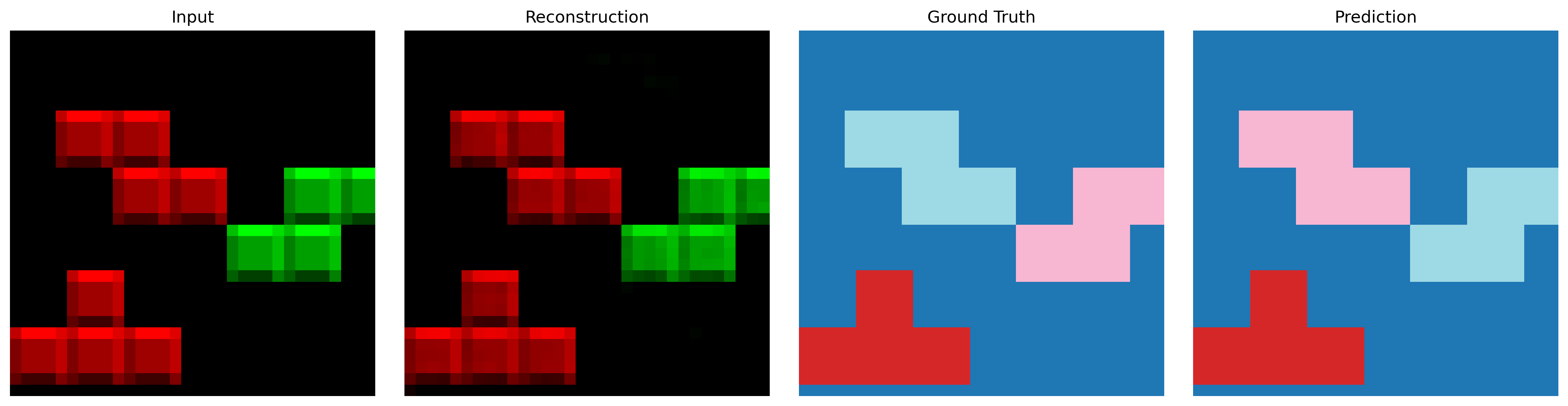} &
 \includegraphics[width=2.5in]{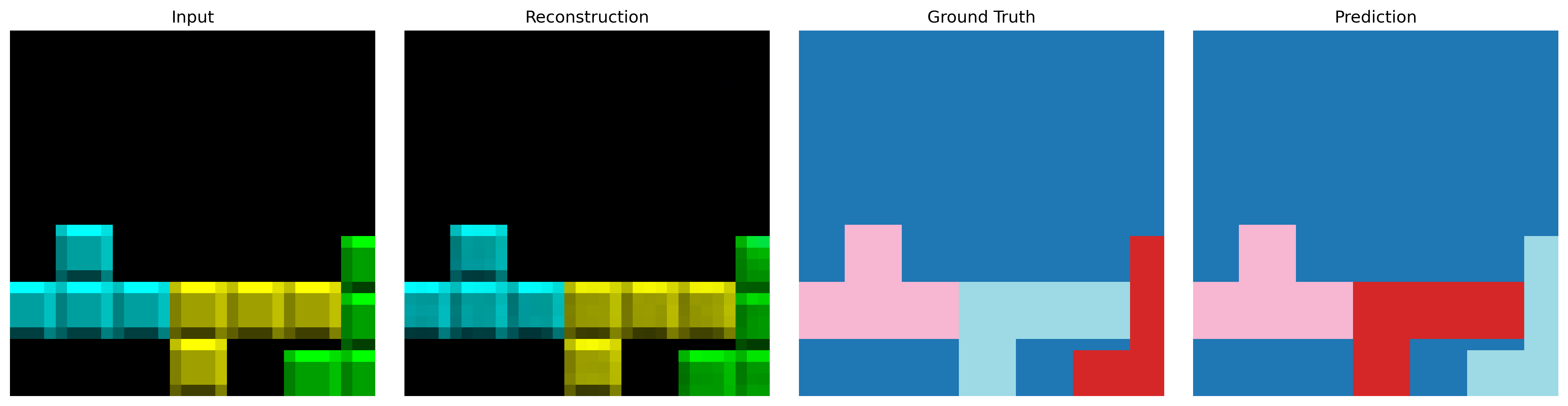} \\
\end{tabular}
\caption{Qualitative results of CORE on the Tetrominoes dataset. From left to right: input image, reconstruction, ground-truth mask, and predicted mask.}
\label{fig:tetr_many}
\end{figure}

\begin{figure}[h]
\centering
{%
  \includegraphics[width=0.7\textwidth]{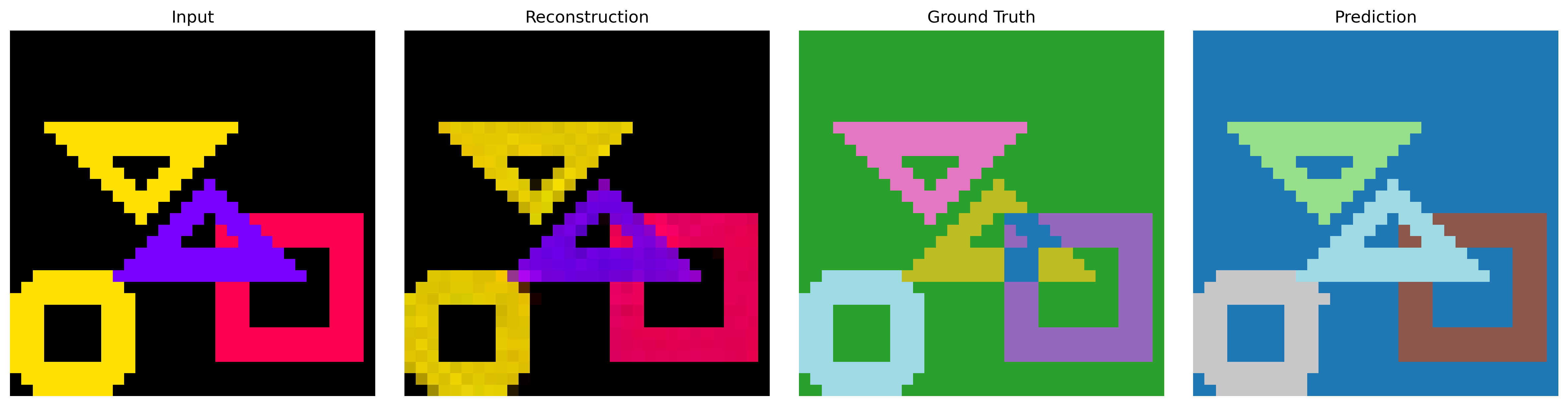}%
  }\par
{%
  \includegraphics[width=0.7\textwidth]{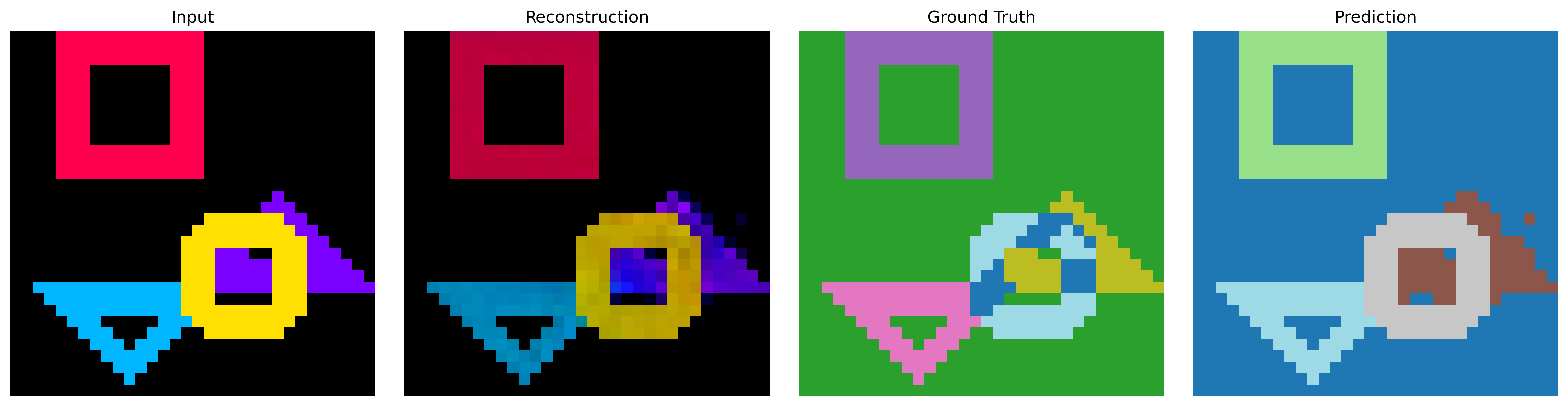}%
 }
{%
  \includegraphics[width=0.7\textwidth]{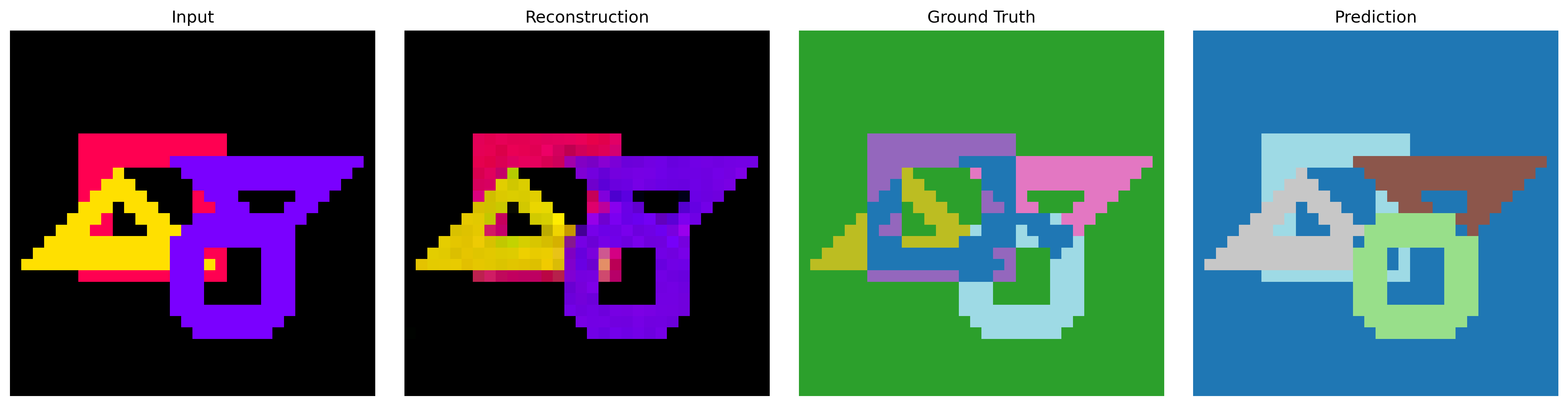}
  }\par
{%
  \includegraphics[width=0.7\textwidth]{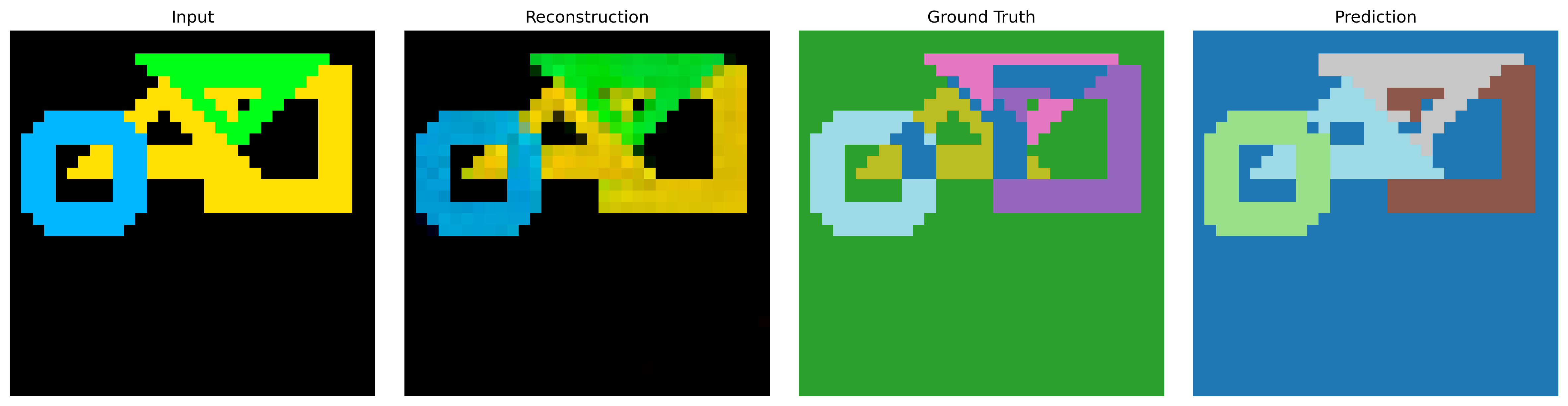}%
 }
 {%
  \includegraphics[width=0.7\textwidth]{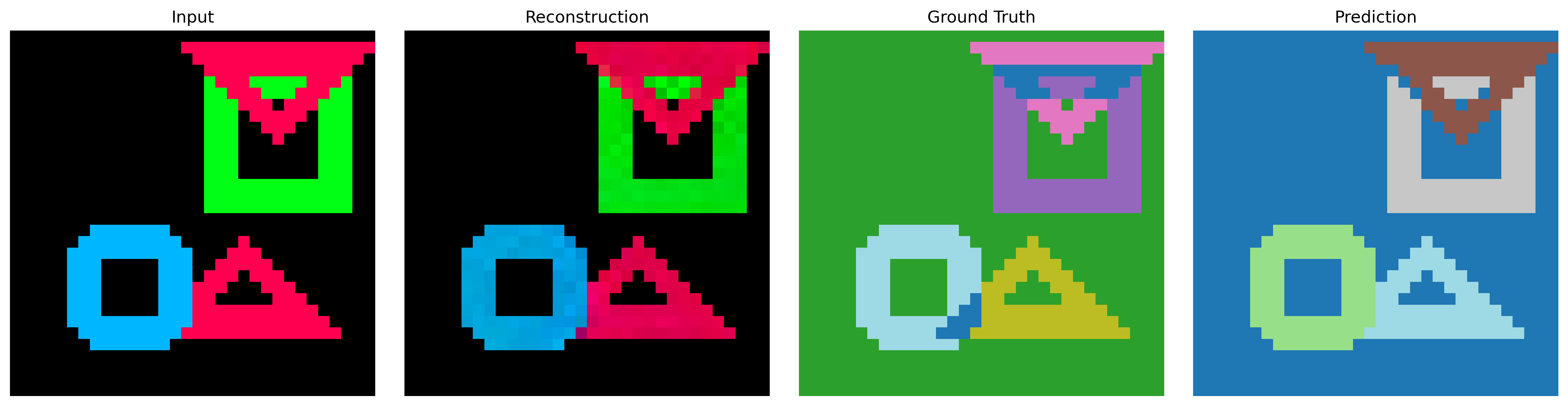}%
  }\par
{%
  \includegraphics[width=0.7\textwidth]{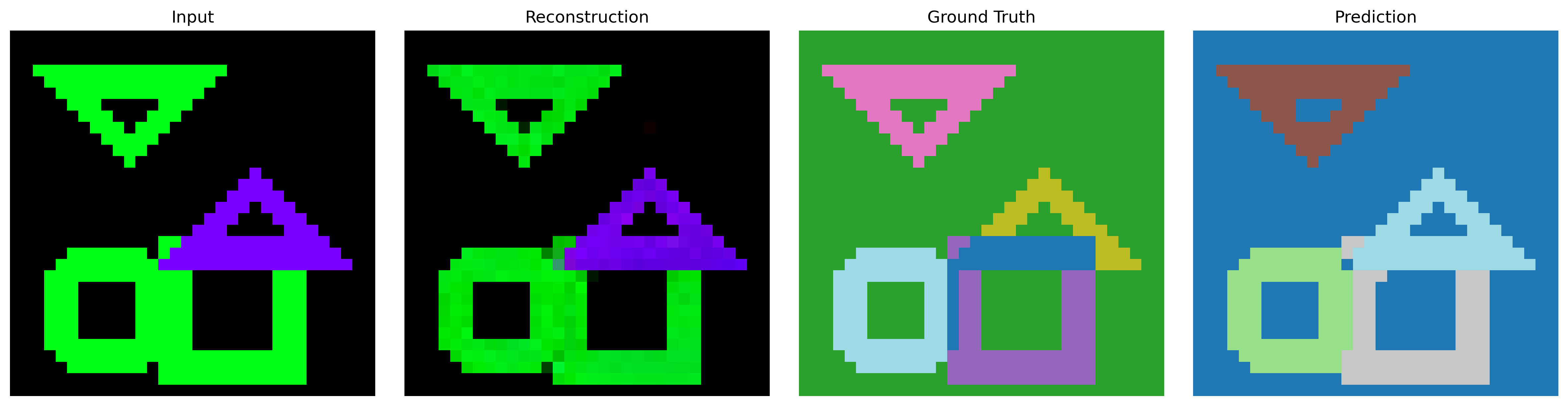}%
 }
{%
  \includegraphics[width=0.7\textwidth]{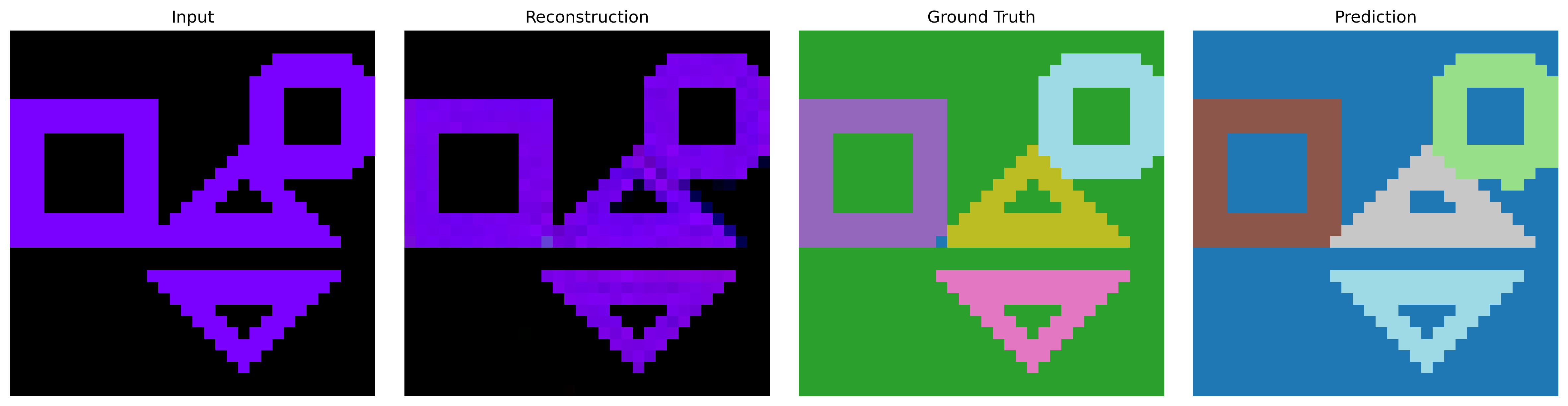}%
 }

\caption{Qualitative results of CORE on the 4Shapes dataset. From left to right: input image, reconstruction, ground-truth mask, and predicted mask.}
\label{fig:4shapes_many}
\end{figure}

\begin{figure}[h]
\centering
{%
  \includegraphics[width=0.6\textwidth]{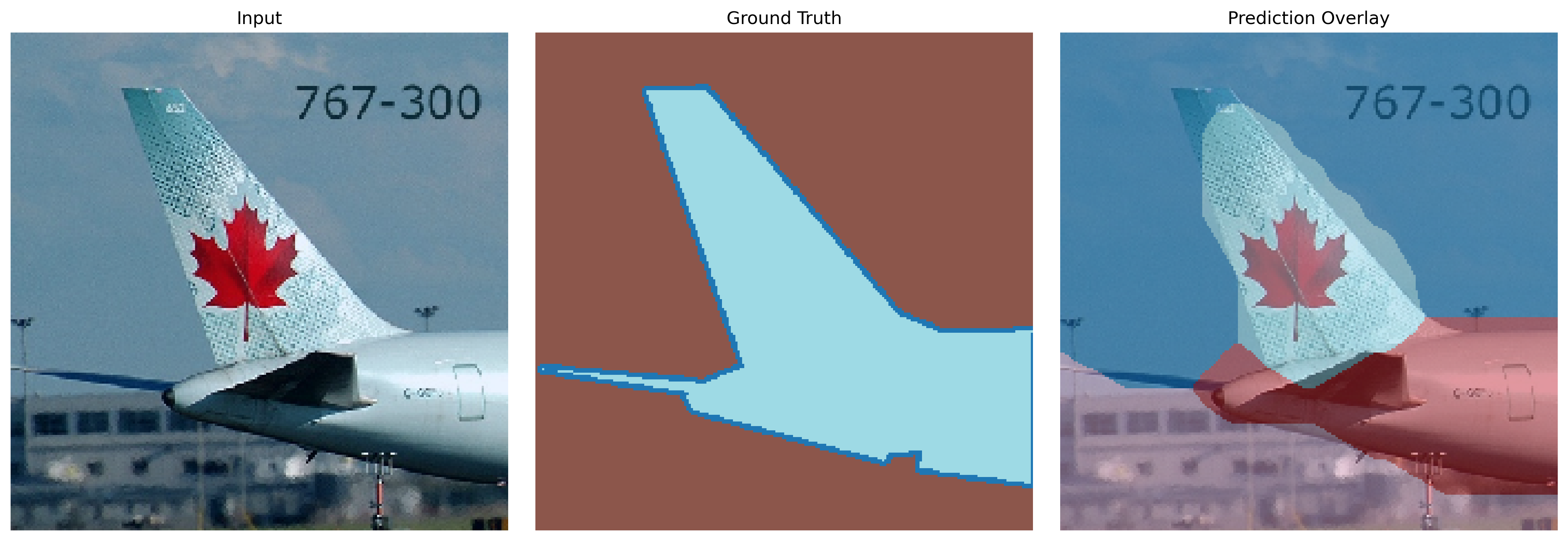}%
  }\par
{%
  \includegraphics[width=0.6\textwidth]{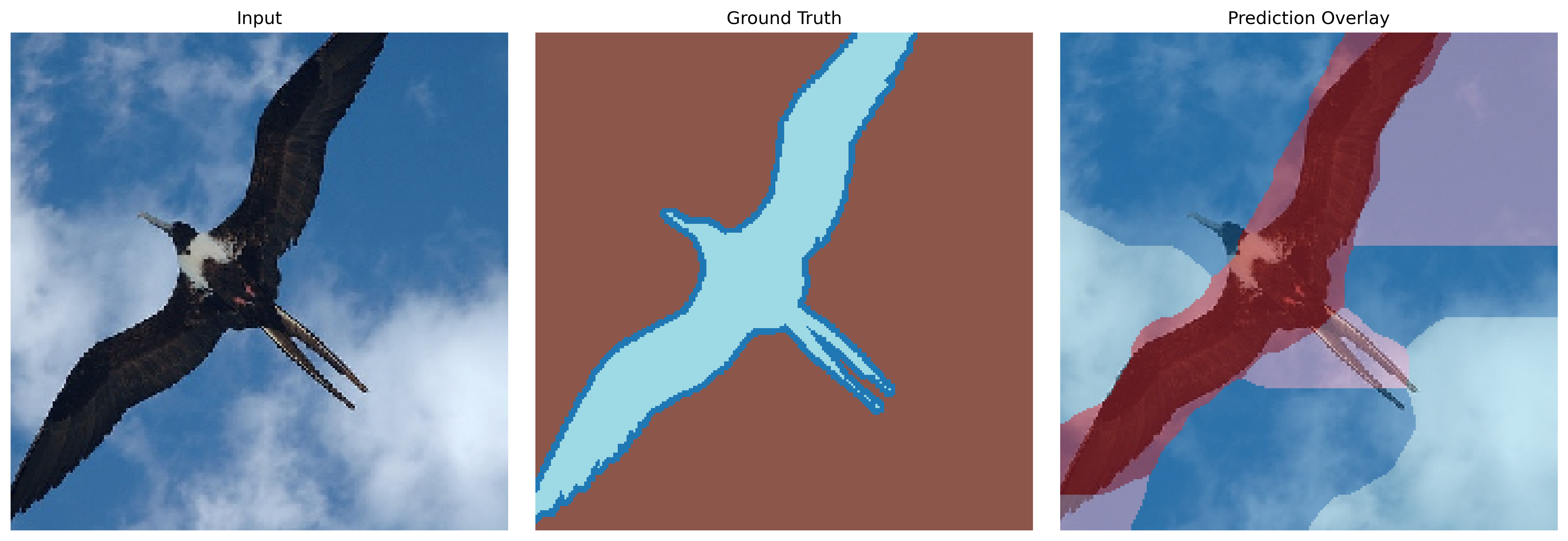}%
 }
{%
  \includegraphics[width=0.6\textwidth]{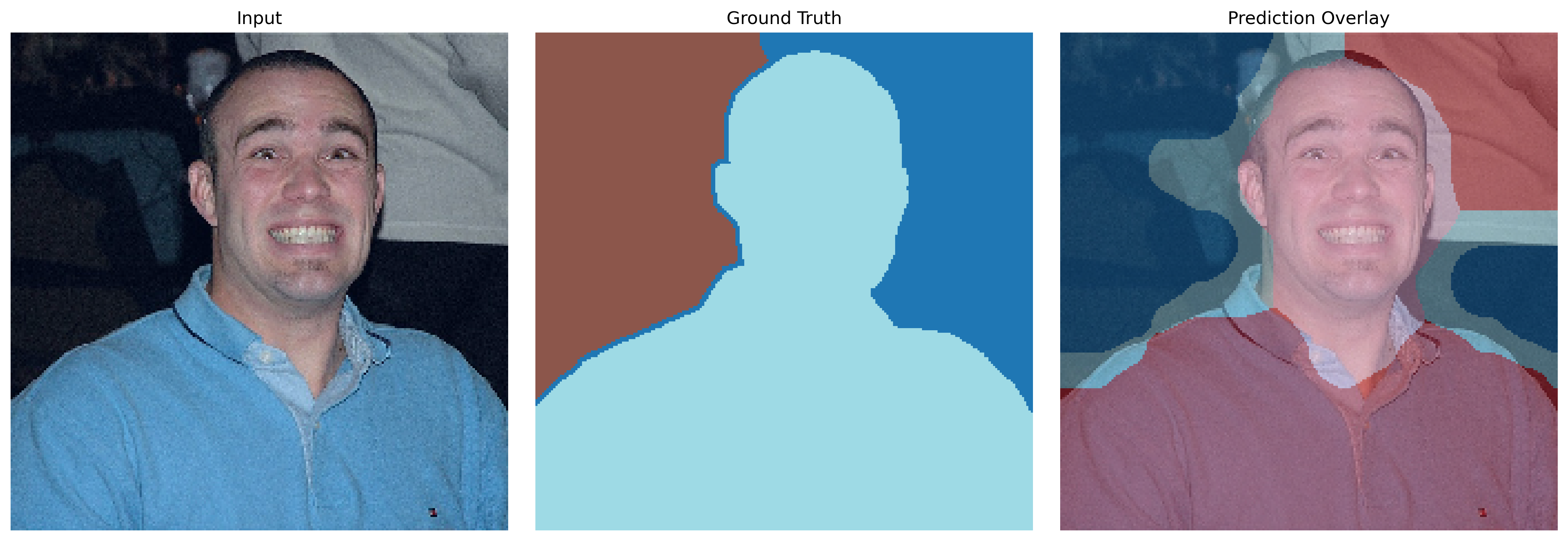}
  }\par
{%
  \includegraphics[width=0.6\textwidth]{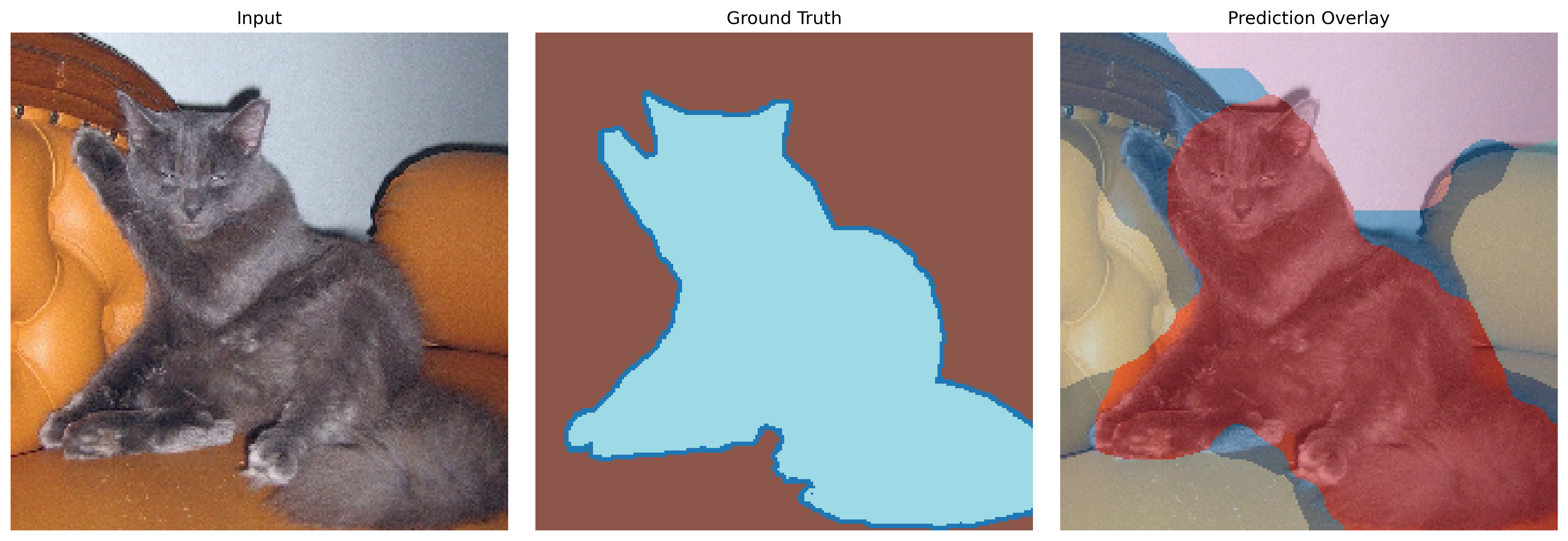}%
 }
 {%
  \includegraphics[width=0.6\textwidth]{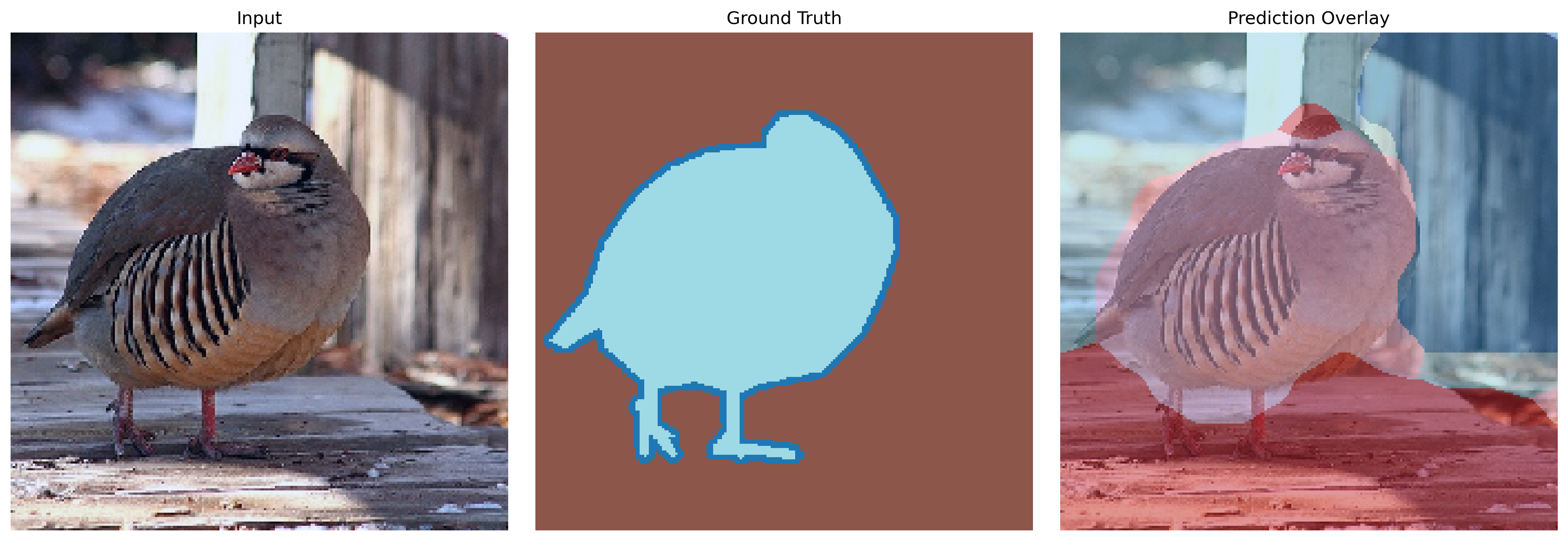}%
  }\par
{%
  \includegraphics[width=0.6\textwidth]{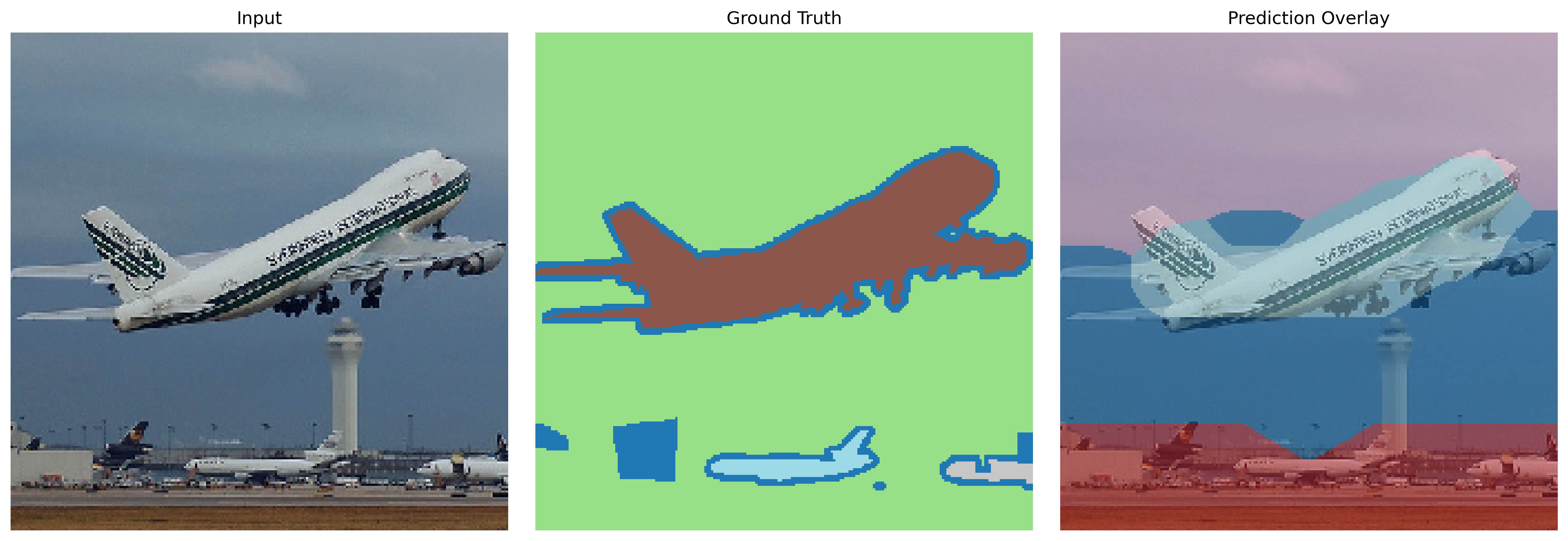}%
 }
{%
  \includegraphics[width=0.6\textwidth]{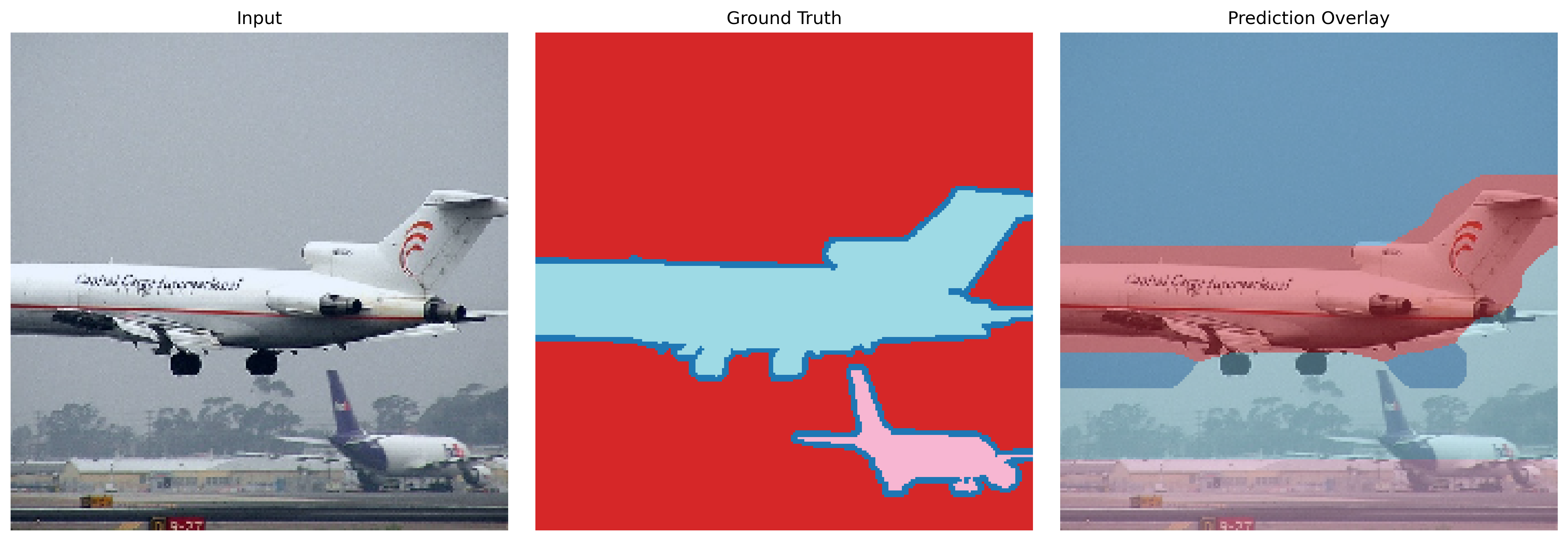}%
 }

\caption{Qualitative results of CORE on the Pascal VOC. From left to right: input image, ground-truth mask, and predicted mask.}
\label{fig:pascal_many}
\end{figure}

We provide additional quantitative results for 4Shapes RGB(-D) in Table~\ref{tab:4shapes_rgb_full}, including RF results when depth masks are used during training. Only with this additional depth supervision does RF approach CORE’s performance; however, CORE still achieves stronger results while using only RGB inputs.
We also provide additional results in Table~\ref{tab:tetr_full}, including evaluation metrics beyond those reported in Table~\ref{tab:tetr} for the Tetrominoes dataset.
Finally, we provide additional qualitative results for CORE on the 4Shapes, Tetrominoes, and Pascal VOC datasets in Figures~\ref{fig:4shapes_many}, \ref{fig:tetr_many}, and \ref{fig:pascal_many}, respectively.

\begin{table}[t]
\caption{
Object discovery results on the 4Shapes RGB dataset, reported as mean \(\pm\) standard deviation over 4 seeds. 
As the number of possible object colors increases, RF performance degrades on the RGB dataset. Providing depth information (RGB-D) mitigates this degradation. In contrast, CORE outperforms RF using RGB alone and maintains stable grouping performance across color-set variants, comparable to RF when depth masks are provided. Best and second-best values are shown as \bestcap{bold} and \secondcap{underlined}, respectively.}
\label{tab:4shapes_rgb_full}
\centering
\renewcommand{\arraystretch}{1.2}
\begin{threeparttable}
\begin{tabular}{l c c c c c}
\hline
\textbf{Model} & \textbf{Dataset} & \textbf{Colors} & \textbf{MSE ↓} & \textbf{ARI-BG ↑} & \textbf{MBO\textsubscript{i} ↑} \\
\hline

\multirow{5}{*}{RF}
& RGB & 1 & \secondsci{1.017}{-5}{\sci{2.541}{-6}} & \valerr{0.982}{0.009} & \valerr{0.874}{0.008} \\
& RGB & 2 & \secondsci{1.597}{-4}{\sci{2.185}{-5}} & \valerr{0.726}{0.035} & \valerr{0.595}{0.015} \\
& RGB & 3 & \secondsci{1.044}{-4}{\sci{6.547}{-6}} & \valerr{0.850}{0.008} & \valerr{0.767}{0.016} \\
& RGB & 4 & \secondsci{1.166}{-4}{\sci{5.985}{-6}} & \valerr{0.682}{0.041} & \valerr{0.626}{0.050} \\
& RGB & 5 & \secondsci{1.653}{-4}{\sci{3.065}{-5}} & \valerr{0.494}{0.056} & \valerr{0.474}{0.031} \\
\hline

\multirow{5}{*}{RF}
& RGB-D & 1 & \bestsci{4.920}{-6}{\sci{5.421}{-7}} & \bestvalerr{1.000}{0.000} & \secondvalerr{0.875}{0.015} \\
& RGB-D & 2 & \bestsci{4.693}{-5}{\sci{2.166}{-5}} & \bestvalerr{0.996}{0.004} & \secondvalerr{0.798}{0.007} \\
& RGB-D & 3 & \bestsci{2.436}{-5}{\sci{3.930}{-6}} & \secondvalerr{0.936}{0.029} & \secondvalerr{0.785}{0.017} \\
& RGB-D & 4 & \bestsci{4.562}{-5}{\sci{4.567}{-6}} & \secondvalerr{0.909}{0.039} & \secondvalerr{0.759}{0.013} \\
& RGB-D & 5 & \bestsci{8.850}{-5}{\sci{3.780}{-5}} & \secondvalerr{0.922}{0.040} & \secondvalerr{0.727}{0.014} \\
\hline

\multirow{5}{*}{CORE}
& RGB & 1 & \sci{2.500}{-4}\pmsem{\sci{7.071}{-5}} & \secondvalerr{0.999}{0.000} & \bestvalerr{0.999}{0.000} \\
& RGB & 2 & \sci{8.000}{-4}\pmsem{\sci{1.001}{-5}} & \secondvalerr{0.996}{0.000} & \bestvalerr{0.981}{0.001} \\
& RGB & 3 & \sci{1.100}{-3}\pmsem{\sci{4.243}{-4}} & \bestvalerr{0.995}{0.002} & \bestvalerr{0.981}{0.005} \\
& RGB & 4 & \sci{2.050}{-3}\pmsem{\sci{9.192}{-4}} & \bestvalerr{0.985}{0.011} & \bestvalerr{0.957}{0.023} \\
& RGB & 5 & \sci{1.700}{-3}\pmsem{\sci{2.828}{-4}} & \bestvalerr{0.991}{0.001} & \bestvalerr{0.966}{0.008} \\
\hline

\end{tabular}
\end{threeparttable}
\end{table}

\newpage

\newpage

\newpage

\begin{table}[t]
\caption{
Object discovery results on the Tetrominoes dataset, reported as mean \(\pm\) standard deviation over 5 seeds. CORE outperforms the other synchrony-based models and achieves performance comparable to Slot Attention~\cite{locatello2020object}. Best and second-best values are shown as \bestcap{bold} and \secondcap{underlined}, respectively. * denotes a run excluded from evaluation as an outlier.}
\label{tab:tetr_full}
\centering
\renewcommand{\arraystretch}{1.2}
\begin{threeparttable}
\resizebox{\linewidth}{!}{%
\begin{tabular}{l c c c c}
\hline
\textbf{Model} & \textbf{MSE ↓} & \textbf{ARI-BG ↑} & \textbf{ARI+BG ↑} & \textbf{MBO\textsubscript{i} ↑} \\
\hline
\multicolumn{5}{l}{\textit{Synchrony-based models}} \\

CAE~\cite{lowe2022complex}
& \sci{4.57}{-2} {\scriptsize$\pm$ \sci{1.08}{-3}}
& \valerr{0.000}{0.000}
& \valerr{0.120}{0.020}
& -- \\

CAE++~\cite{stanic2023contrastive}
& \bestsci{5.07}{-5}{\sci{2.80}{-5}}
& \valerr{0.780}{0.070}
& \valerr{0.840}{0.010}
& -- \\

CtCAE~\cite{stanic2023contrastive}
& \secondsci{9.73}{-5}{\sci{4.64}{-5}}
& \valerr{0.840}{0.090}
& \valerr{0.850}{0.010}
& -- \\

SynCx~\cite{gopalakrishnan2024recurrent}
& --
& \valerr{0.890}{0.001}
& --
& -- \\

RF~\cite{lowe2023rotating}
& \sci{3.00}{-4} {\scriptsize$\pm$ \sci{1.00}{-4}}
& \valerr{0.420}{0.060}
& --
& \valerr{0.440}{0.080} \\

ItrSA~\cite{miyato2024artificial}
& --
& \valerr{0.858}{0.008}
& --
& \valerr{0.549}{0.003} \\

AKOrN~\cite{miyato2024artificial}
& --
& \valerr{0.886}{0.002}
& --
& \secondvalerr{0.564}{0.009} \\

AKOrN {\scriptsize\textit{+ up-tiling}}~\cite{miyato2024artificial}
& --
& \secondvalerr{0.931}{0.003}
& --
& \valerr{0.563}{0.000} \\

CORE
& \sci{4.00}{-4} {\scriptsize$\pm$ \sci{1.00}{-4}}
& \bestvalerr{0.976}{0.001}
& \bestvalerr{0.989}{0.002}
& \bestvalerr{0.976}{0.002} \\

\hline
\multicolumn{5}{l}{\textit{Slot-based model}} \\

Slot Attention~\cite{locatello2020object}
& --
& \bestvalerr{0.990}{0.200}*
& --
& -- \\

\hline
\end{tabular}%
}
\end{threeparttable}
\end{table}


\clearpage

\end{document}